\documentclass[preprint,review,3p,times]{elsarticle}

\usepackage{hyperref}
\usepackage{caption}
\usepackage{amssymb}
\usepackage{amsmath}
\usepackage{enumitem}
\usepackage{algorithm}
\usepackage{algpseudocode}
\usepackage{array}
\usepackage{booktabs} 
\usepackage{subcaption}
\usepackage{multirow} 



\begin{document}

\begin{frontmatter}



\title{Think-Before-Draw: Decomposing Emotion Semantics \& Fine-Grained Controllable Expressive Talking Head Generation}


\author{Hanlei Shi} 
\author{Leyuan Qu}
\author{Yu Liu}
\author{Di Gao}
\author{Yuhua Zheng\corref{cor1}}
\author{Taihao Li\corref{cor1}}
\affiliation{
            addressline={Hangzhou Institute for Advanced Study, University of Chinese Academy of Sciences}, 
            city={Hangzhou},
            country={China}}
\cortext[cor1]{Corresponding author}
\begin{abstract}
Emotional talking-head generation has emerged as a pivotal research area at the intersection of computer vision and multimodal artificial intelligence, with its core value lying in enhancing human-computer interaction through immersive and empathetic engagement.
With the advancement of multimodal large language models, the driving signals for emotional talking-head generation has shifted from audio and video to more flexible text. However, current text-driven methods rely on predefined discrete emotion label texts, oversimplifying the dynamic complexity of real facial muscle movements and thus failing to achieve natural emotional expressiveness.
This study proposes the Think-Before-Draw framework to address two key challenges: (1) In-depth semantic parsing of emotions—by innovatively introducing Chain-of-Thought (CoT), abstract emotion labels are transformed into physiologically grounded facial muscle movement descriptions, enabling the mapping from high-level semantics to actionable motion features; and (2) Fine-grained expressiveness optimization—inspired by artists' portrait painting process, a progressive guidance denoising strategy is proposed, employing a ``global emotion localization—local muscle control" mechanism to refine micro-expression dynamics in generated videos.
Our experiments demonstrate that our approach achieves state-of-the-art performance on widely-used benchmarks, including MEAD and HDTF. Additionally, we collected a set of portrait images to evaluate our model's zero-shot generation capability.

\end{abstract}

\begin{graphicalabstract}
\end{graphicalabstract}

\begin{highlights}
\item We introduce Think-Before-Draw, a novel framework to achieve fine-grained emotional talking-head video generation under text guidance.
\item CoT technology maps emotional semantics to facial movements via multi-step analysis, transforming abstract labels into physiologically grounded descriptions.
\item Progressive guidance denoising enhances video expressiveness via global-local emotion-to-muscle control.
\item Analysis shows our framework excels in emotional expressiveness, motion naturalness, and user control for interactive virtual humans.

\end{highlights}


\end{frontmatter}



\section{Introduction}
With the rapid advancement of the metaverse, virtual digital humans, and intelligent interaction technologies~\cite{metaverse,metaversesurvey}, emotional talking-head generation has become a key research focus in the interdisciplinary field of computer vision and multimodal artificial intelligence~\cite{zhen2023human,zhou2024survey}. Emotional expression serves as a core element of human-computer interaction, profoundly shaping user immersion, trust, and empathetic engagement. Consequently, creating realistic emotionally expressive talking heads continues to pose a significant challenge in this field. 
The task of emotional talking-head generation is to synthesize a lifelike talking face from a single static image and driving signals, ensuring lip synchronization, natural head movements, and nuanced emotional expression. This technology demonstrates significant potential in various fields, such as digital assistants, film-making, and vitrual video conferences.

In recent years, research on emotional talking-head generation has attracted growing attention~\cite{ji2021Audio-Driven,ji2022EAMM,ma2023styletalk,zhang2023sadtalker,peng2023emotalk,gan2023EAT,liu2025moee}. These methods either accomplish expression transfer by assigning expressions frame-by-frame from input video templates, or enable expression manipulation using predefined emotion labels. 
With the rapid advancement of multimodal large language models (MLLMs)~\cite{survey-MLLM,qwen2}, researchers have begun exploring more flexible driving approaches, particularly using natural language text to control emotional expressions~\cite{gan2023EAT,fg-emotalk,expclip,pan2024expressive}. However, these text-driven approaches often rely on simplistic label-to-text mappings, where discrete emotion labels are directly used as textual descriptors. This crude ``label-to-expression" mapping frequently results in unnatural and stiff facial animations—exaggerated expressions that mismatch the character’s identity or even violate biomechanical constraints. Facial expressions arise from complex muscle coordination rather than simple linear mappings~\cite{ekman1978facial}. In simple terms, single emotion labels fail to capture the nuanced interplay of facial muscles in real emotional expressions. 

\begin{figure}[!t]
\centering
\includegraphics[width=4in]{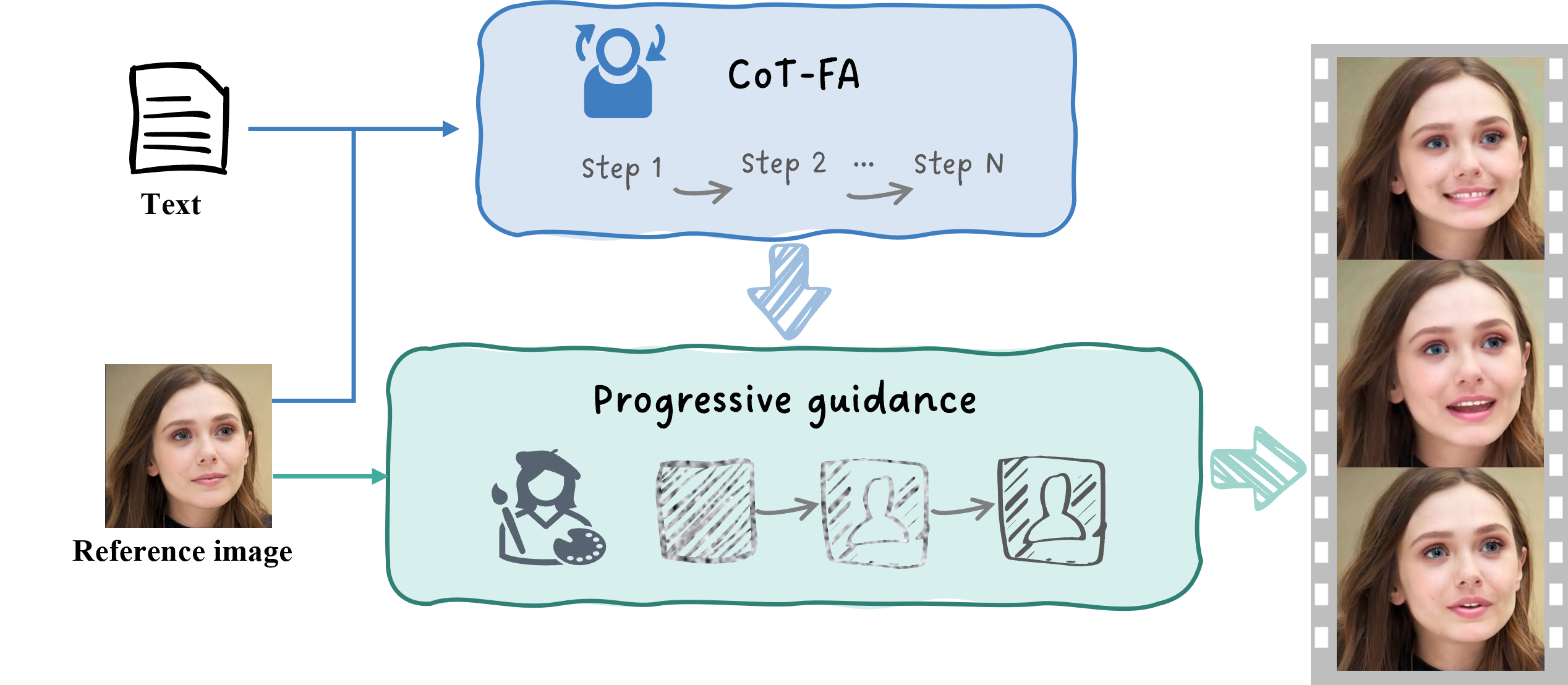}
\captionsetup{font=normalsize}
\caption{Illustration of our proposed framework. Our framework primarily consists of two components: Chain-of-Thought Facial Animation (CoT-FA) and progressive guidance generation. Given a reference image and text input, the reference image and text are processed by the CoT-FA module to generate video prompts. The progressive guidance component then synthesizes the final video by combining these prompts with the reference image.}
\label{fig_1}
\end{figure}

To achieve fine-grained, text-guided emotional talking-head video generation, we address the following two key challenges: (1) investigating how to perform in-depth semantic parsing of a single abstract emotion label, transforming its high-level semantic information into actionable facial motion features for video generation, and (2) thoroughly exploring how to optimize the detailed expressiveness of generated videos through fine-grained control strategies. To address the challenge mentioned above, we present Think-Before-Draw, a novel framework that integrates Chain-of-Thought (CoT)~\cite{wei2022CoT,zhang2022automatic} with a progressive guidance denoising strategy to emotion-aware talking head generation as shown in \autoref{fig_1}.

Firstly, inspired by the mechanisms of facial expression generation, facial expressions fundamentally arise from the fine-grained coordinated movements of multiple facial muscle groups~\cite{ekman1978facial,jarlier2011muscle,2005facialanalysis,duchenne1990mechanism}. We propose to incorporate knowledge of facial muscle kinematics into the field of expression analysis. And then, the CoT technique which explicitly models intermediate analytical processes, effectively guides models to emulate the progressive thinking characteristics of human cognition, thereby significantly enhancing their analytical capabilities for complex problems. 
Building upon facial expressions foundation and CoT technique, we presents an innovative Chain-of-Thought facial animation (CoT-FA) module for emotion labeling that systematically bridges the gap between facial muscle movements and expressive semantics. By integrating the Facial Action Coding System (FACS)~\cite{ekman1978facial} standard with physiological knowledge~\cite{jarlier2011muscle,duchenne1990mechanism} about core muscle groups underlying dominant emotional states, our approach establishes a fine-grained bidirectional mapping between Facial Action Units (AUs)~\cite{ekman1978facial} and their corresponding semantic interpretations, significantly enhancing the explainability of facial expression analysis.

Secondly, to achieve more natural and vivid facial expression synthesis, inspired by artists' portrait painting process~\cite{faigin2012artist,loomis2021drawing} which progresses from holistic composition to fine details, we develop a progressive guidance denoising strategy that employs hierarchical conditioning during the denoising process. Building upon our CoT-FA module which generates multi-granularity text prompts, we implement a hierarchical conditioning mechanism operating in two complementary phases: (1) coarse-grained prompts establish the emotional foundation in early denoising steps, while (2) fine-grained prompts enable nuanced expression control in later stages.
This multi-scale control mechanism demonstrates exceptional capability in capturing and reproducing the subtle dynamic characteristics of human facial expressions, offering a effective solution for fine-grained emotional expression manipulation. 
In summary, the main contributions of our research are:

\begin{itemize}
\item We introduce Think-Before-Draw (TBD), a novel framework that integrates CoT with a progressive guidance denoising strategy, which achieving fine-grained emotional talking-head video generation under text guidance.
\end{itemize}
\begin{itemize}
\item To addressing the challenge of mapping emotional semantics to visual expression, the innovative introduction of CoT technology transforms abstract emotional labels into physiologically grounded facial muscle movement descriptions through multi-step analysis. 
\end{itemize}
\begin{itemize}
\item To enhancing the emotional expressiveness and naturalness of generated videos, a progressive guidance denoising strategy is proposed, employing a ``global emotion localization—local muscle control" mechanism. 
\end{itemize}
\begin{itemize}
\item Through quantitative and qualitative analysis, this study validates the proposed framework's significant advantages in emotional expressiveness, motion naturalness, and user control convenience, offering new technical insights for emotionally interactive virtual humans.
\end{itemize}

\section{Related Work}
\subsection{Emotion-aware talking head generation}
In the field of emotion-aware talking head generation, researchers focus on modifying or regenerating facial expressions while preserving lip synchronization and natural facial movements~\cite{xu2024hallo,echomimic}. Current research primarily revolves around three driving approaches, each exhibiting distinct advantages and limitations.
Audio-driven methods generate facial animations by extracting emotional features from speech signals~\cite{zhang2023sadtalker,sonic}. While these techniques achieve lip-sync accuracy, they suffer from inherent limitations in precise emotion expression: the many-to-one mapping between acoustic features and emotional states restricts the reliability of emotion inference.
Reference video-driven approaches~\cite{ma2023styletalk,ji2022EAMM,Edtalk} capture emotional style features from reference videos. However, their practical application faces two key constraints: the high cost of acquiring high-quality reference videos and the risk of identity feature leakage during style transfer.
In contrast, label-driven or text-driven paradigms~\cite{gan2023EAT,fg-emotalk,expclip,pan2024expressive} provide users with a more intuitive editing interface, enabling precise emotion control through natural language descriptions, thereby significantly lowering the usability barrier. Nevertheless, existing text-driven methods inherently struggle with the semantic gap in cross-modal mapping from text descriptions to facial actions.
\subsection{Chain-of-Thought}
Chain-of-Thought~\cite{wei2022CoT} represents a groundbreaking reasoning paradigm whose essence lies in guiding large language models to adopt progressive problem-solving strategies, systematically decomposing complex tasks into interpretable intermediate reasoning steps. This explicit reasoning process generation mechanism not only significantly enhances the model's logical reasoning capabilities but also provides transparent decision-making foundations~\cite{wei2022emergent,prystawski2023think}. In multimodal research, MM-CoT~\cite{MM-COT} pioneered the extension of this paradigm to visual-language interaction tasks, with its innovative two-stage fine-tuning architecture (reasoning generation phase and answer inference phase) achieving deep synergistic processing of cross-modal information. Visual CoT~\cite{visual-cot} further optimized multimodal model performance by organically integrating dynamic visual attention mechanisms with chain-of-thought reasoning.

As research progresses, CoT technology has demonstrated its application value in generative visual content creation. Innovative frameworks like Show-o~\cite{xie2024show-O} have achieved unified processing of autoregressive modeling and discrete diffusion models, with this flexible architectural design enabling adaptation to diverse multimodal input-output requirements. The model exhibits exceptional versatility across various tasks including visual question answering, text-to-image generation, text-guided image editing, and hybrid modality generation. GoT~\cite{guo2025GoT} conducted the first comprehensive exploration of CoT reasoning strategies in autoregressive image generation. Dysen-VDM~\cite{fei2024dysen} innovatively employed large language models for motion semantic parsing and transformed them into Dynamic Scene Graph (DSG)~\cite{dsg} representations, providing a novel technical pathway for controllable video generation. These cutting-edge achievements not only validate the universality of the CoT mechanism in cross-modal generation tasks but also offer important insights for the theoretical framework construction and methodological design of this study.

\section{Method}
\begin{figure*}[!t]
\centering
\includegraphics[width=6 in]{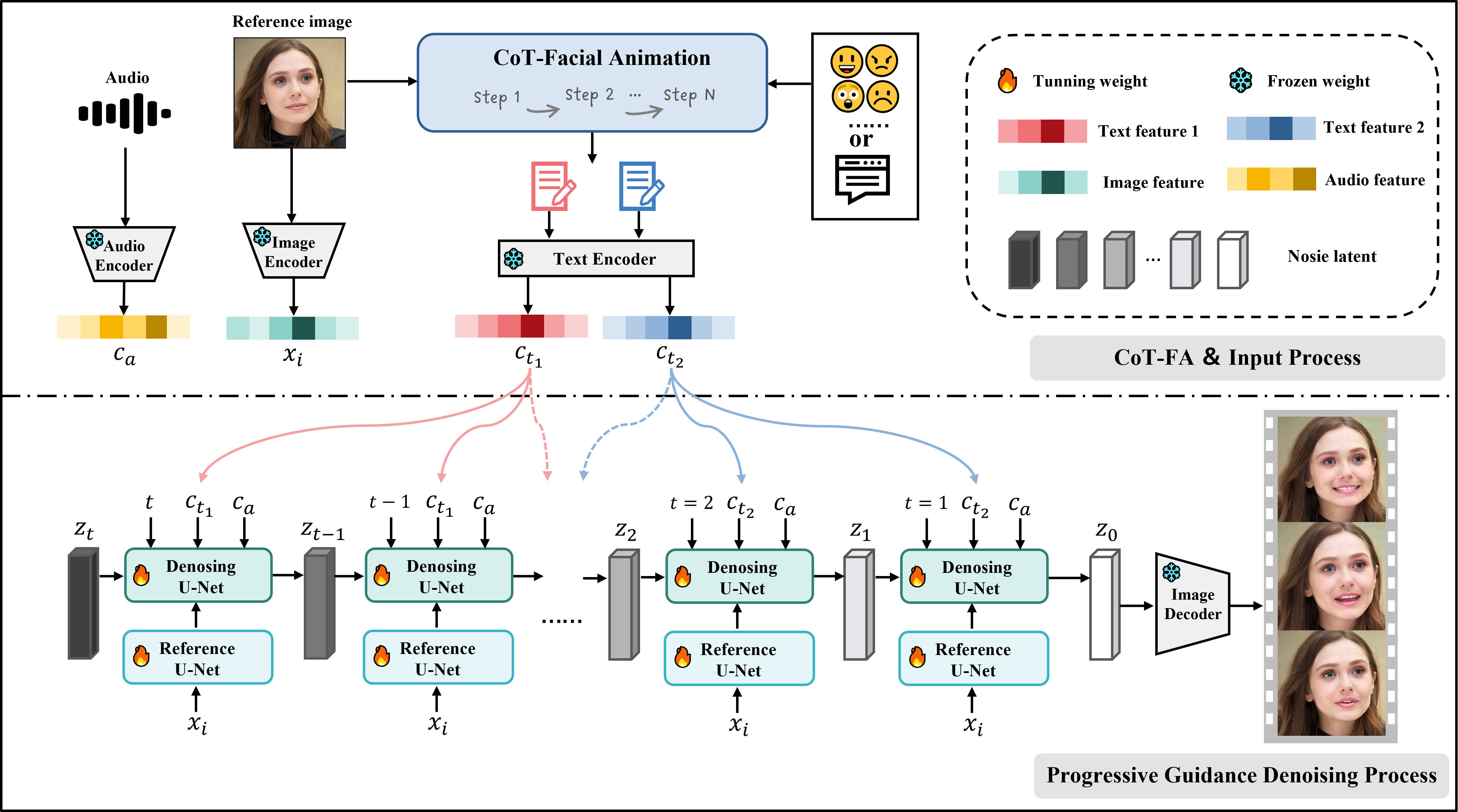}
\captionsetup{font=normalsize}
\caption{\textbf{The overview of Think-Before-Draw.}  In the CoT-FA \& Input Processing stage, the CoT-FA module processes the reference image and text, generating multi-level descriptions encoded by text encoder into text embeddings $c_{t_1}, c_{t_2}$. The image encoder extracts image features $x_i$, while the audio encoder extracts audio features $c_a$.
In the progressive guidance denoising processing stage, we employed a diffusion-based generative model. During the denoising phase of the diffusion model, the image features $x_i$ are input as latent vectors into the reference U-Net, which are then fed into the Denoising U-Net. Additionally, audio features $c_a$ and text conditions ($c_{t_1}$ for global guidance in early steps; $c_{t_2}$ for details in late steps) are progressively fed to the Denoising U-Net.} 
\label{fig_2}
\end{figure*}

\subsection{Overall framework}
The pipeline of our method is illustrated in \autoref{fig_2}. Given an input reference image, audio, and text containing emotion-related keywords, we employ Wav2Vec~\cite{wav2vec} as the audio encoder to extract audio features and VAE encoder~\cite{VAE} as the image encoder to extract reference image features. For text processing (detailed implementation in \ref{subsec2}), the input is first fed into the CoT-FA module to generate multi-level textual descriptions, which are then passed to the CLIP~\cite{clip} text encoder. In detail, when a user-provided text instruction or emotion label is processed by the CoT-FA module, it yields both coarse-grained and fine-grained textual descriptions. These descriptions are encoded into text embeddings, denoted as $c_{t_1}$ and $c_{t_2}$, serving as hierarchical text-guided conditions. During the generation phase, our framework builds upon recent diffusion-based talking head generation methods~\cite{hallo,echomimic}, with modifications to certain modules in the Denoising U-Net~\cite{stablediffusion,Animatediff} to enable text-driven control (architecture details in \ref{subsec3}). For the denoising stage (full implementation of the progressive guidance strategy is elaborated in \ref{subsec4}), we propose a progressively guided strategy: specifically, the coarse-grained embedding $c_{t_1}$ is used as the global guidance in the early denoising steps, while the fine-grained embedding $c_{t_2}$ is applied in the later stages, achieving precise text control from holistic to detailed levels.

\begin{figure}[!t]
\centering
\includegraphics[width=3.5in]{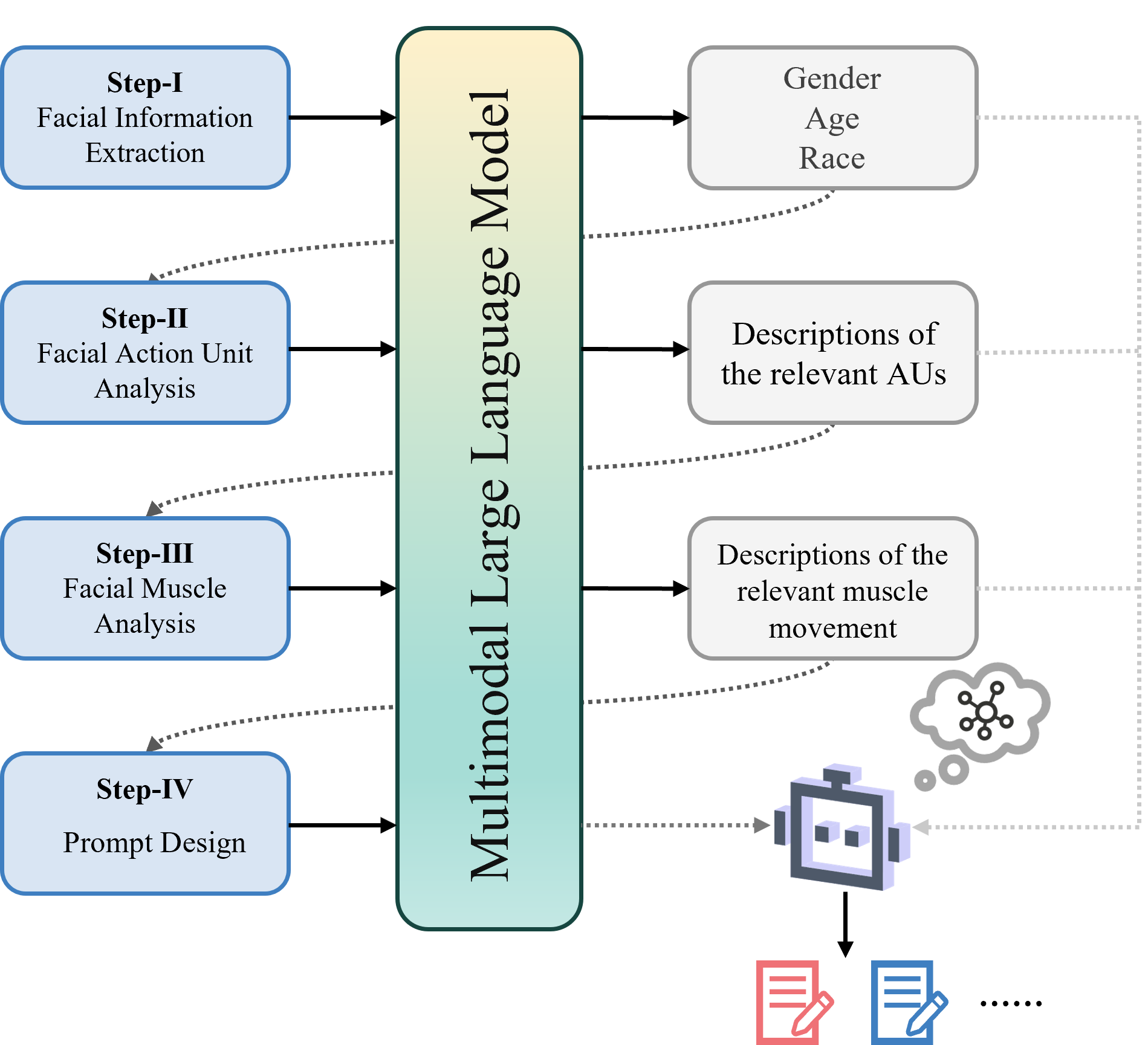}
\captionsetup{font=normalsize}
\caption{The overview of Chain-of-Thought Facial Animation. We establish a four-tier progressive processing pipeline: (1) facial information extraction, (2) facial action unit analysis, (3) facial muscle analysis, and (4) prompt design.}
\label{fig_3}
\end{figure}

\begin{figure*}[!t]
\centering
\includegraphics[width=5.5 in]{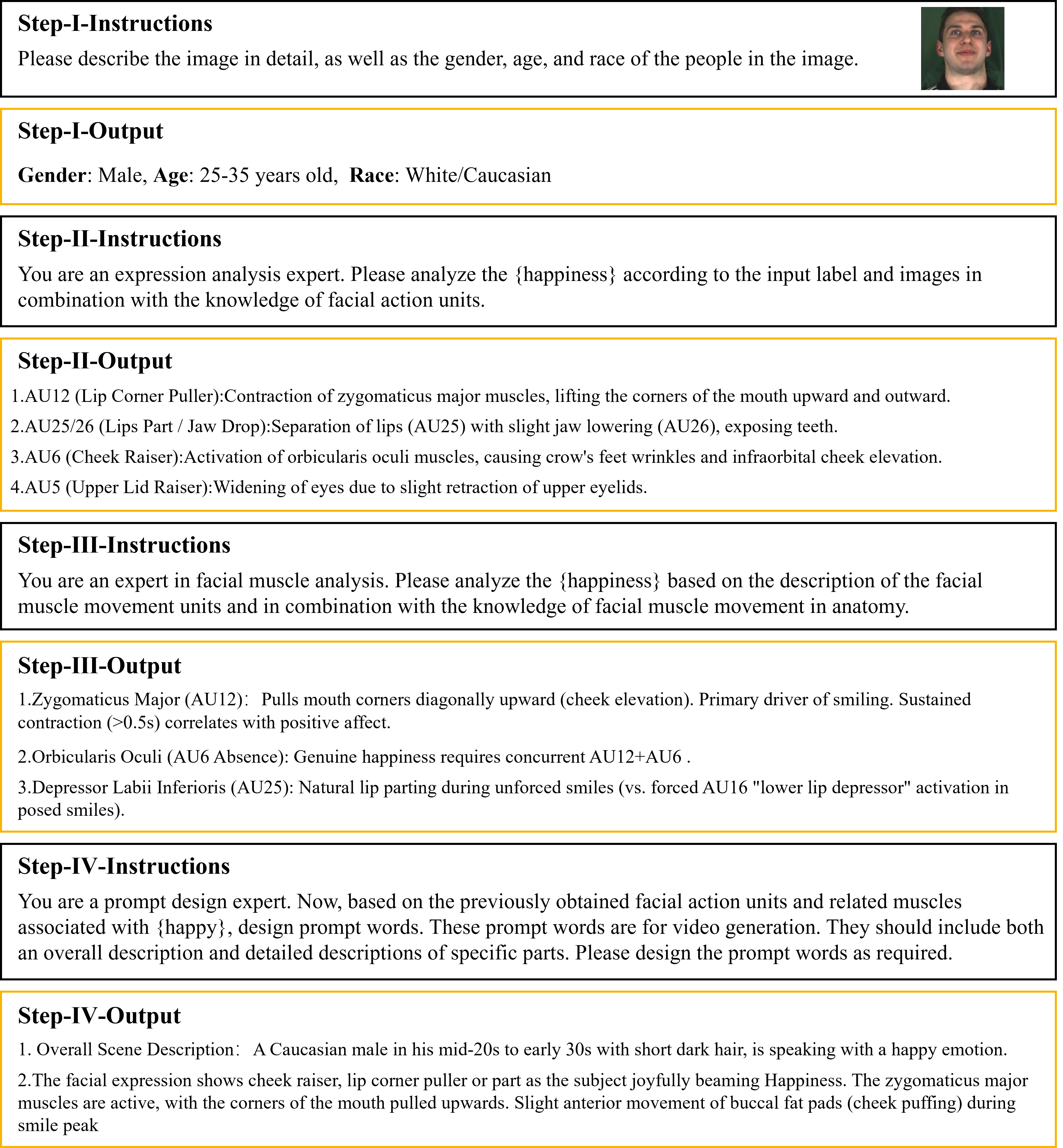}
\captionsetup{font=normalsize}
\caption{Illustrative example of CoT-FA} 
\label{fig_4}
\end{figure*}

\subsection{Chain-of-Thought Facial Animation}
\label{subsec2}
As previously discussed, using conventional emotion labels as drivers suffers from limitations in expressiveness and abstraction.  Inspired by facial expression generation mechanisms, we recognize that facial expressions fundamentally arise from fine-grained coordination of multiple facial muscle groups. 
To address the challenge of mapping emotional semantics to visual expression, this study integrates facial muscle kinematics into expression analysis and employs CoT technology to guide the model in simulating human cognition through step-by-step analysis, from character attributes to coarse-grained expressions and finally to fine-grained facial muscle descriptions.

We innovatively propose a CoT-based facial animation labeling module that systematically bridges facial muscle movements with expressive semantics. Grounded in the FACS~\cite{ekman1978facial} standard, this mechanism deconstructs emotion labels into specific muscle movement descriptions by identifying core muscle groups associated with dominant emotional states and incorporating anatomical studies to generate text descriptions that comply with real muscle movement patterns.
As illustrated in \autoref{fig_3}, this framework fully leverages the advantages of state-of-the-art multimodal large language models (particularly Qwen2-VL~\cite{qwen2} in cross-modal understanding and knowledge reasoning, establishing a four-tier progressive processing pipeline : (1) facial information extraction, (2) facial action unit analysis, (3) facial muscle analysis, and (4) prompt design.
Through this four-stage analytical pipeline, the multimodal large language models systematically deconstruct emotion labels by integrating identity attributes from reference images, Facial Action Unit semantics based on FACS standards, and biomechanical knowledge of muscle kinematics to generate anatomically grounded prompts for emotionally expressive talking head generation. Each data entry undergoes manual review to filter out anomalous descriptions.

\textbf{Step-I: facial information extraction}

In the initial processing stage, we employ multimodal interaction to acquire fundamental identity characteristics from reference images, including key parameters such as age range, gender characteristics, and ethnic attributes. This information provides crucial contextual foundation for subsequent refined analysis.

\textbf{Step-II: facial action unit analysis}

The second phase adopts a professional analysis method based on the FACS~\cite{ekman1978facial} system, where we guide the multimodal model to function as a facial expression analysis expert. When input with specific emotion labels and reference images, the model precisely identifies key AUs activated in that emotional state based on FACS expertise, and returns textual descriptions of the relevant AUs. This standardized coding system ensures scientific rigor and reproducibility in emotion representation.

\textbf{Step-III: facial muscle analysis}

In this more advanced analytical stage, the model shifts to a facial muscle kinetics specialist perspective. This phase requires the model to not only understand the muscle movement mechanisms corresponding to each AUs, but also comprehensively consider factors including facial anatomical characteristics, muscle synergies, and individual variations. Notably, as facial expressions result from combined physiological and sociocultural factors, the analysis must respect both the physical principles of muscle movement and the sociocultural dimensions of expression.

\textbf{Step-IV: prompt design}

In the final stage, building upon previous analysis results, the model functions as a prompt design expert to generate multi-level descriptive texts encompassing both holistic and detailed features. 

Taking ``happiness" as an example (\autoref{fig_4}), the system outputs not only holistic descriptions like ``A Caucasian male in his mid-20s to early 30s with short dark hair, is speaking with a happy emotion", but also professional-level details such as ``The facial expression shows cheek raiser, lip corner puller or part as the subject joyfully beaming Happiness. The zygomaticus major muscles are active, with the corners of the mouth pulled upwards. Slight anterior movement of buccal fat pads (cheek puffing) during smile peak". This dual-level description approach, combining macroscopic characteristics with microscopic details, provides precise biological guidance and rich semantic information for subsequent visual content generation. Subsequently, each data entry will go through manual inspection to filter out abnormal descriptions.

\subsection{Enhanced denoising U-Net}
\label{subsec3}
\begin{figure}[!t]
\centering
\includegraphics[width=4 in]{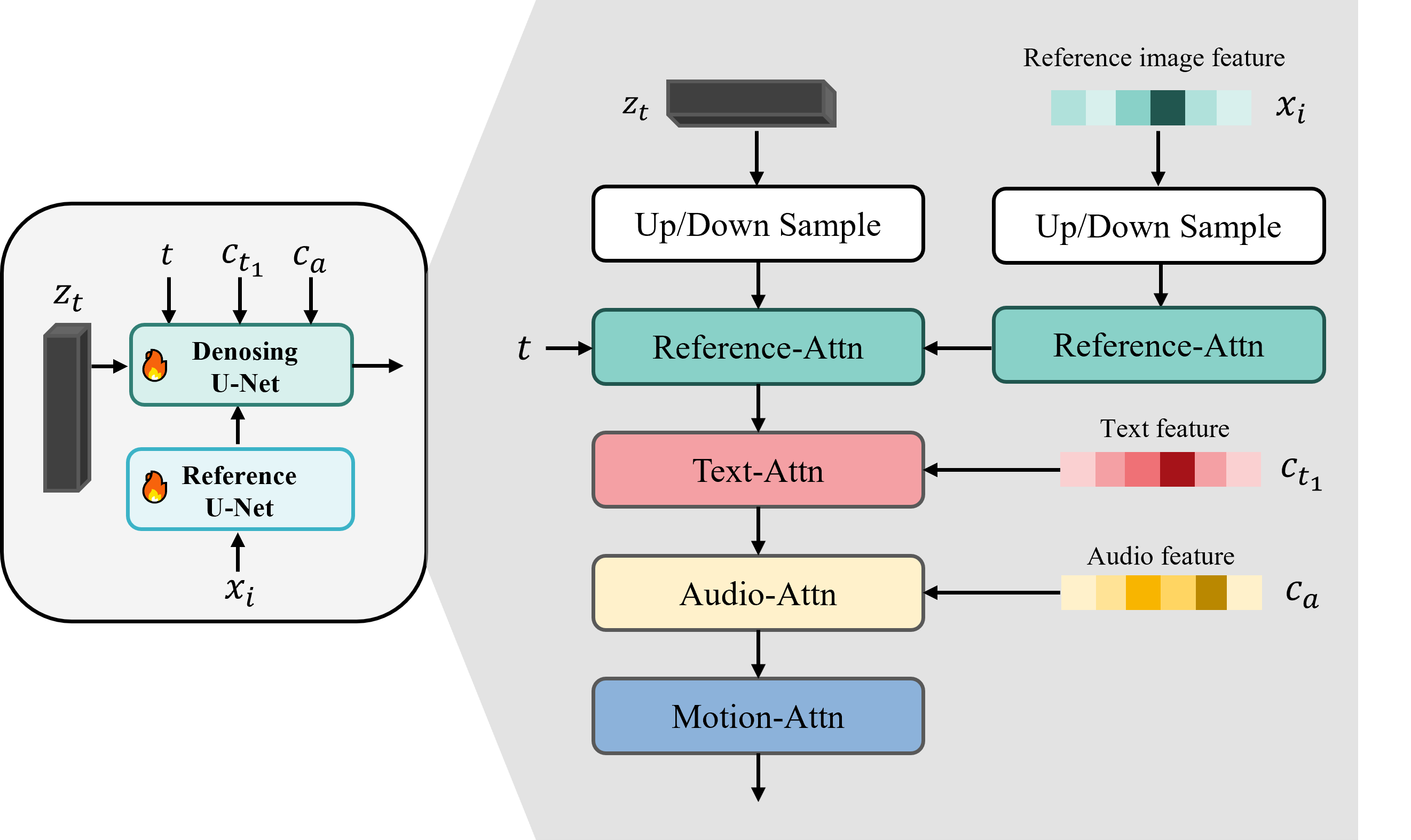}
\captionsetup{font=normalsize}
\caption{Illustration of either a CrossAttnDownBlock or CrossAttnUpBlock within the denoising U-Net architecture. The noisy latent variable $z_t$ first undergoes processing through either an upsampling or downsampling module. Subsequently, $z_t$ flows into the Reference-Attn module, where it performs attention operations with reference image features to capture critical information correlations.
The processed $z_t$ then sequentially enters both the Text-Attn and Audio-Attn modules, executing attention computations with textual and audio features respectively, thereby integrating multimodal information. Finally, the data passes through the Motion-Attn module to complete the entire processing pipeline, achieving deep fusion and feature extraction of multimodal information.}
\label{fig_5}
\end{figure}

Our foundational framework builds upon the denoising U-Net architecture~\cite{stablediffusion,Animatediff,hallo}, which achieving high-quality talking face video generation through multimodal feature fusion. 
As illustrated in the \autoref{fig_5}.

\begin{itemize}
\item \textbf{Reference attention layer} establishes cross-frame correspondence with reference images to maintain identity consistency. 
\end{itemize}
\begin{itemize}
\item \textbf{Text attention layer} aligns CLIP-encoded text features with visual features for fine-grained emotion control. 
\end{itemize}
\begin{itemize}
\item \textbf{Audio attention layer} focuses on spatiotemporal alignment between speech features and facial movements to ensure precise lip synchronization.
\end{itemize}
\begin{itemize}
\item \textbf{Temporal attention layer} employs inter-frame self-attention to capture dynamic video sequence patterns and ensure motion coherence~\cite{Animatediff}. This hierarchical attention architecture enables the model to simultaneously handle complex spatial and temporal relationships.
\end{itemize}

Our denoising U-Net builds upon the standard Stable Diffusion v1.5 (SDv1.5)~\cite{stablediffusion} architecture with key improvements through multi-level attention mechanisms. SD integrates principles from the Denoising Diffusion Probabilistic Model (DDPM)~\cite{DDPM} or its variant, the Denoising Diffusion Implicit Model (DDIM)~\cite{DDIM}, introducing a strategic element of gaussian noise $\epsilon$ to the latent representation $z_0$, yielding a temporally indexed noisy latent state $z_t$ at step t. The objective function guiding the denoising process during training is formulated as follows:
\begin{equation}
\mathcal{L}=\mathbb{E}_{t,c_{t},c_{a},z_t,\epsilon}[\lVert \epsilon-\epsilon_\theta(z_t,t,c_{t},c_{a}) \rVert^2_2]
\end{equation}

Here, $c_t$ signifies the text features extracted from the input text prompt utilizing the CLIP text encoder. Meanwhile, $c_a$ stands for the audio features derived from the input audio prompt through the utilization of the Wav2Vec audio encoder. Within the Stable Diffusion framework, the estimation of the noise $\epsilon$ is accomplished by a customized UNet architecture. This UNet model has been augmented with across-attention mechanism, allowing for the effective integration of text features $c_t$ and audio features $c_a$ with the latent representation $z_t$.
The cross attention operation is formulated as:
\begin{equation}
CrossAttn(z_t,c_{embed})=softmax(QK^T\sqrt{d})V
\end{equation}

where $Q=W_QZ_t, K=W_kC$ and $V=W_VC$ are the queries; $W_Q, W_k$ and $W_v$ are learnable projection matrices; and $d_k$ is the dimensionality of keys. $c_{embed}$ represents either $c_t$ or $c_a$.

\subsection{Progressive guidance denoising strategy}
\label{subsec4}
To achieve more natural and vivid facial expression generation, inspired by artists' portrait painting process which progresses from holistic composition to fine details, we proposes an innovative progressive guidance denoising strategy that hierarchically optimizes the denoising generation process~\cite{faigin2012artist,loomis2021drawing}. When creating portraits, artists typically follow a holistic-to-detailed workflow: they first outline the figure's basic structure, then precisely position facial feature proportions, subsequently refine facial details progressively, and finally perform overall harmonization and adjustments. This process emphasizes establishing fundamental relationships first to avoid premature focus on isolated details, thereby ensuring anatomical accuracy and compositional harmony.
The core concept of the progressive guidance denoising strategy involves sequentially introducing multi-level guidance information—from macro to micro—during different denoising stages: the initial phase establishes the overall emotional tone through high-level semantic concepts (e.g., happiness, anger), while subsequent phases incorporate fine-grained muscle movement descriptions based on the FACS (e.g., zygomaticus major contraction, corrugator supercilii activation).
\begin{figure}[!t]
\centering
\includegraphics[width=3.5 in]{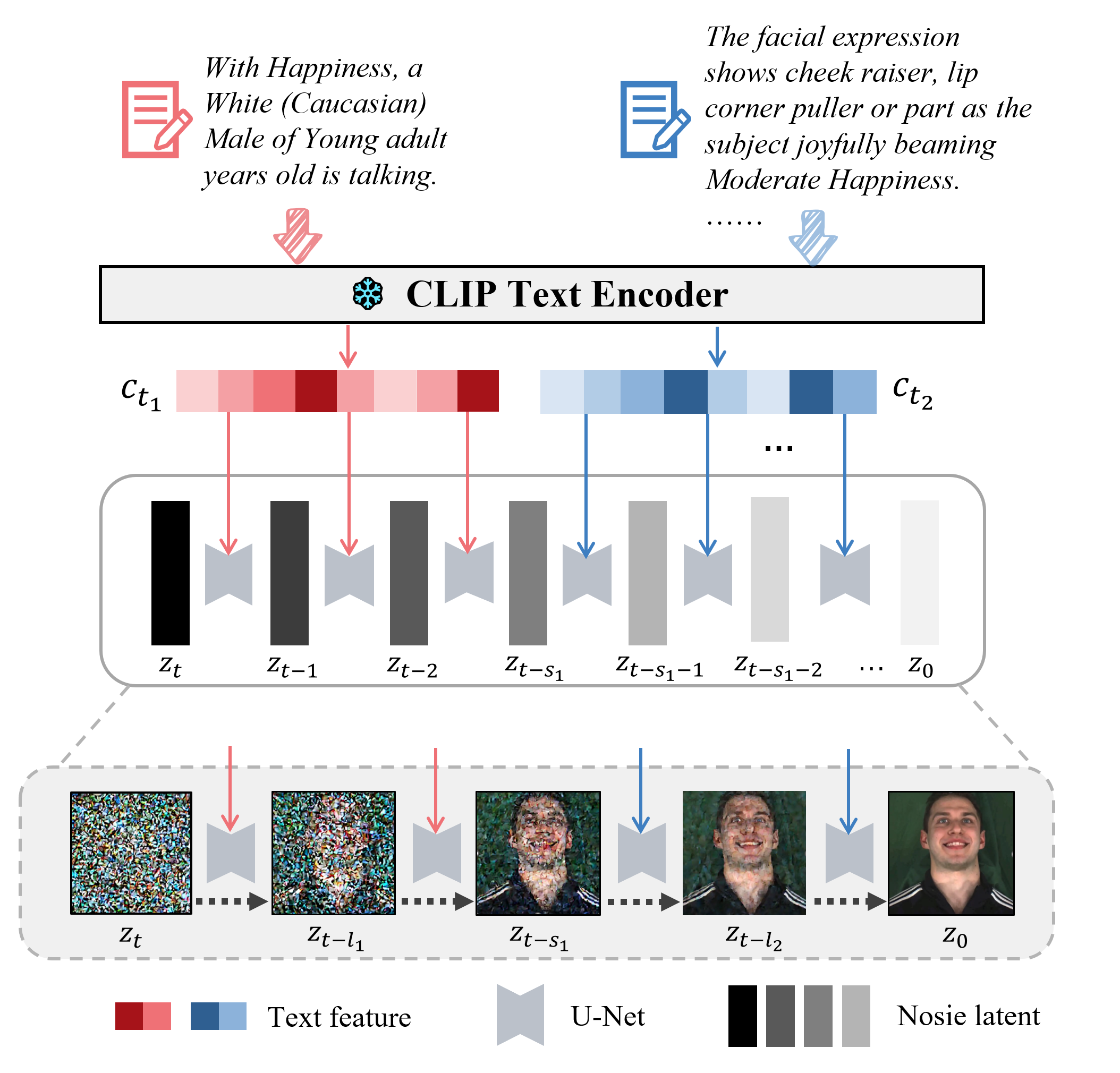}
\captionsetup{font=normalsize}
\caption{Illustration of progressive guidance denoising strategy. In this example, there are only two-stage guidance (k=2) and the number of total guidance steps is $t$ with $s_1=t * 0.4$. Dark gray to light gray embeddings represent the denoising process in the latent space; the lighter the color, the less noise the image latent embedding contains.($0\le{l_1}\le{s_1}$, $s_1\le{l_2}\le{t}$)}
\label{fig_6}
\end{figure}

Given a latent diffusion model $\mathcal{M}$ with $t$ sampling steps and an ordered sequence of $k$ semantic-guided text prompts $\lbrace{P_1, P_2,...,P_k} \rbrace$, the video generation process proceeds under their sequential guidance. First, a Gaussian-noised video is encoded into an initial latent vector $Z_t$. The $k$ text prompts are then separately embedded into vectors $\lbrace{e_1, e_2,...,e_k} \rbrace$ via a pretrained text encoder. Let $\lbrace{s_1, s_2,...,s_k} \rbrace$ denote the sampling steps allocated to each text-guidance stage, where $\sum{s_i}=t$. The initial latent vector $Z_t$  is progressively transformed into the final latent vector $Z_0$ through $t$ denoising steps, guided sequentially by $\lbrace{e_1, e_2,...,e_k} \rbrace$. In the denoising U-Net text-attention module, noise prediction $n_t$ is computed based on the current timestep $t$, input latent vector $Z_t$, and current-stage text embedding $e_t$. Finally, $Z_0$ is decoded into the target video.

\begin{algorithm}[t]
\caption{Alpha Value Calculation}
\label{alg:alpha}
\begin{algorithmic}[1]
\Require Current timestep $t$, stage start $t_s$, stage end $t_e$, maximum $\alpha_{max}$, minimum $\alpha_{min}$
\Ensure Alpha value $\alpha \in [\alpha_{min}, \alpha_{max}]$
\State Compute normalized position: $p \gets (t - t_s)/(t_e - t_s)$
\State Calculate alpha: $\alpha \gets \alpha_{max} - p \cdot (\alpha_{max} - \alpha_{min})$
\State \Return $\alpha$
\end{algorithmic}
\end{algorithm}

\begin{algorithm}[t]
\caption{Text Embedding Adjustment}
\label{alg:text_embed}
\begin{algorithmic}[1]
\Require Timesteps $\mathbf{t} \in \mathbb{R}^{bs}$, embeddings $\mathbf{text}_{s1}, \mathbf{text}_{s2}$, stage end $t_{end}$, total timesteps $T$
\Ensure Adjusted embedding $\mathbf{text}$
\State Initialize $\mathbf{text} \gets \mathbf{0}$ \Comment{Same shape as $\mathbf{text}_{s1}$}
\For{each sample $i$ in batch}
    \State $t \gets \mathbf{t}[i]$
    \If{$t < t_{end}$}
        \State $\alpha \gets \text{calculate\_alpha}(t, 0, t_{end}, 1.0, 0.5)$
        \State $\mathbf{text}[i] \gets \alpha \cdot \mathbf{text}_{s1}[i] + (1-\alpha) \cdot \mathbf{text}_{s2}[i]$
    \Else
        \State $\alpha \gets \text{calculate\_alpha}(t, t_{end}, T, 1.0, 0.5)$
        \State $\mathbf{text}[i] \gets \alpha \cdot \mathbf{text}_{s2}[i] + (1-\alpha) \cdot \mathbf{text}_{s1}[i]$
    \EndIf
\EndFor
\State \Return $\mathbf{text}$
\end{algorithmic}
\end{algorithm}
Taking k=2 as an example to illustrate the concrete implementation method. As illustrated in the \autoref{fig_6}. After processing the user input text through the Chain-of-Thought Facial Animation, the generated text prompt is divided into two parts: during the first stage of the denoising process ($t\in\left[ 0,s_1 \right]$), the macro-level emotional description ``A Caucasian male in his mid-20s to early 30s with short dark hair, is speaking with a happy emotion" was employed as the guiding prompt ($c_{t_1}$), focusing on controlling the character's overall emotional expression. Upon entering the second stage ($l\in\left[ s_1,t \right]$), it switches to the micro-level detail description ``The facial expression shows cheek raiser, lip corner puller or part as the subject joyfully beaming Moderate Happiness...". This prompt ($c_{t_2}$) achieves refined control over expression details through precise facial action coding, realizing progressive feature guidance from global to local levels. 

To ensure training coherence, we design a progressive text-guided fusion mechanism with the following key innovations: at stage transitions, a hybrid strategy blends two adjacent text conditions to avoid abrupt guidance shifts. The blending weight $\alpha$ is dynamically adjusted based on the relative position of the current timestep within its stage (rather than using fixed values). This design ensures natural transitions between guidance stages while enhancing the model’s ability to fuse multi-level semantic information, significantly improving the coherence and naturalness of emotional expressions in generated videos.

\section{Experiments}
\subsection{Experimental setup}

\textbf{Implementation details.} This study conducts experiments covering both training and inference phases, performed on high-performance computing hardware equipped with 8 NVIDIA L20 GPUs. The training process consists of two stages, each comprising 30,000 steps, executed with a batch size of 2 while processing video data at a resolution of 512×512 pixels. In the second training phase, each iteration produces 14 video frames. Throughout both phases, the learning rate remains constant at 1e-5. To expedite convergence, the motion module is initialized with pretrained weights from the Animatediff~\cite{Animatediff} model.

\textbf{Datasets.}
As mentioned earlier, this study employs a curated version of the MEAD~\cite{wang2020mead} dataset and the  HDTF~\cite{HDTF} dataset for training . For the MEAD dataset, 80\% of the data is used for training, while the remaining 20\% is reserved for testing and evaluation. For the HDTF dataset, we selected 20 subjects, with 10 video clips randomly chosen for each subject, each lasting approximately 5 to 10 seconds. Since HDTF does not include emotion annotations, we assigned neutral emotion labels to its textual descriptions.

\textbf{Comparison setting.}
We compare our method with: (a) emotion-agnostic talking face generation methods: MakeItTalk~\cite{makelttalk}, SadTalker~\cite{zhang2023sadtalker}, PC-AVS~\cite{PCAVS}. (b) emotional talking face generation methods: EAMM~\cite{ji2022EAMM}, StyleTalk~\cite{ma2023styletalk}, EAT~\cite{gan2023EAT}. 

\textbf{Evaluation metrics.}
To comprehensively evaluate the quality of the generated talking head videos, we employ a combination of widely-used objective metrics that assess different aspects of visual fidelity and perceptual quality. Specifically, we adopt the Peak Signal-to-Noise Ratio (PSNR)~\cite{PSNR} and Structural Similarity Index (SSIM)~\cite{SSIM} to measure pixel-level and structural similarity between the generated videos and the ground-truth frames. Additionally, we utilize the Fréchet Inception Distance (FID~\cite{Seitzer2020FID}) to evaluate the realism and feature-level distribution alignment of the synthesized images with real samples. To further quantify the sharpness and perceptual blurriness of the generated frames, we incorporate the Cumulative Probability of Blur Detection (CPBD~\cite{CPBD_Alt}), which provides a robust estimation of image clarity based on human visual perception. For assessing the accuracy of lip synchronization, we leverage the confidence score from SyncNet (SyncConf~\cite{sync}), a well-established metric that evaluates the temporal alignment between speech audio and corresponding lip movements.

\subsection{Quantitative Results}

The quantitative results are presented in \autoref{tab_1}. On the MEAD dataset, our method achieves the best performance across most metrics. It demonstrates superior and stable performance across image quality, as well as motion synchronization. On the HDTF dataset, our method demonstrates superior performance across most evaluation metrics, except for SyncConf. This discrepancy primarily arises from our text-driven approach: since all HDTF samples contain neutral emotional expressions, the driving texts exhibit inherent uniformity, leading to marginally inferior lip-sync accuracy. It should be noted that this observation is also linked to SadTalker's synthesis mechanism. Specifically, SadTalker employs video-driven synthesis and utilizes Wav2Lip's pre-trained weights for lip movements, while Wav2Lip's SyncNet discriminator—pre-trained on a large-scale dataset~\cite{wav2lip-data}—prioritizes achieving higher synchronization confidence.

\begin{table}[ht]
\centering
\captionsetup{font=normalsize}
\caption{Quantitative comparisons with state-of-the-art methods on MEAD/HDTF dataset.}
\begin{tabular}{l|ccccc} 
\toprule
\multirow{2}{*}{Methods} & \multicolumn{5}{c}{MEAD / HDTF} \\ 
\cline{2-6} 
& PSNR$\uparrow$ & SSIM$\uparrow$ & FID$\downarrow$ & CPBD$\uparrow$ & Sync$_{\text{conf}}$$\uparrow$ \\
\midrule
MakeItTalk & 19.43 / 22.00 & 0.61 / 0.71 & 37.92 / 28.24 & 0.11 / 0.25 & 2.21 / 3.16 \\
SadTalker & 19.04 / 21.70 & 0.69 / 0.70 & 39.31 / 22.06 & 0.16 / 0.24 & 2.76 / \textbf{4.35} \\
PC-AVS & 16.12 / 22.61 & 0.49 / 0.50 & 38.68 / 26.04 & 0.07 / 0.13 & 2.18 / 2.70 \\
EAMM & 18.87 / 19.87 & 0.61 / 0.63 & 31.27 / 41.2 & 0.08 / 0.14 & 1.76 / 2.54 \\
StyleTalk & 21.60 / 21.32 & 0.84 / 0.69 & 24.78 / 17.05 & 0.16 / 0.30 & 3.55 / 3.16 \\
EAT & 20.13 / 22.07 & 0.65 / 0.59 & 21.46 / 29.76 & 0.15 / 0.26 & 2.16 / 3.95 \\
\midrule
\textbf{TBD(Ours)} & \textbf{28.64} / \textbf{23.28} & \textbf{0.88} / \textbf{0.74} & \textbf{17.28} / \textbf{16.17} & \textbf{0.24} / \textbf{0.31} & \textbf{4.71} / 4.10 \\
\bottomrule
\end{tabular}
\label{tab_1}
\end{table}

\begin{figure}[!t]
\centering
\captionsetup{font=normalsize}
\includegraphics[width=6.5 in]{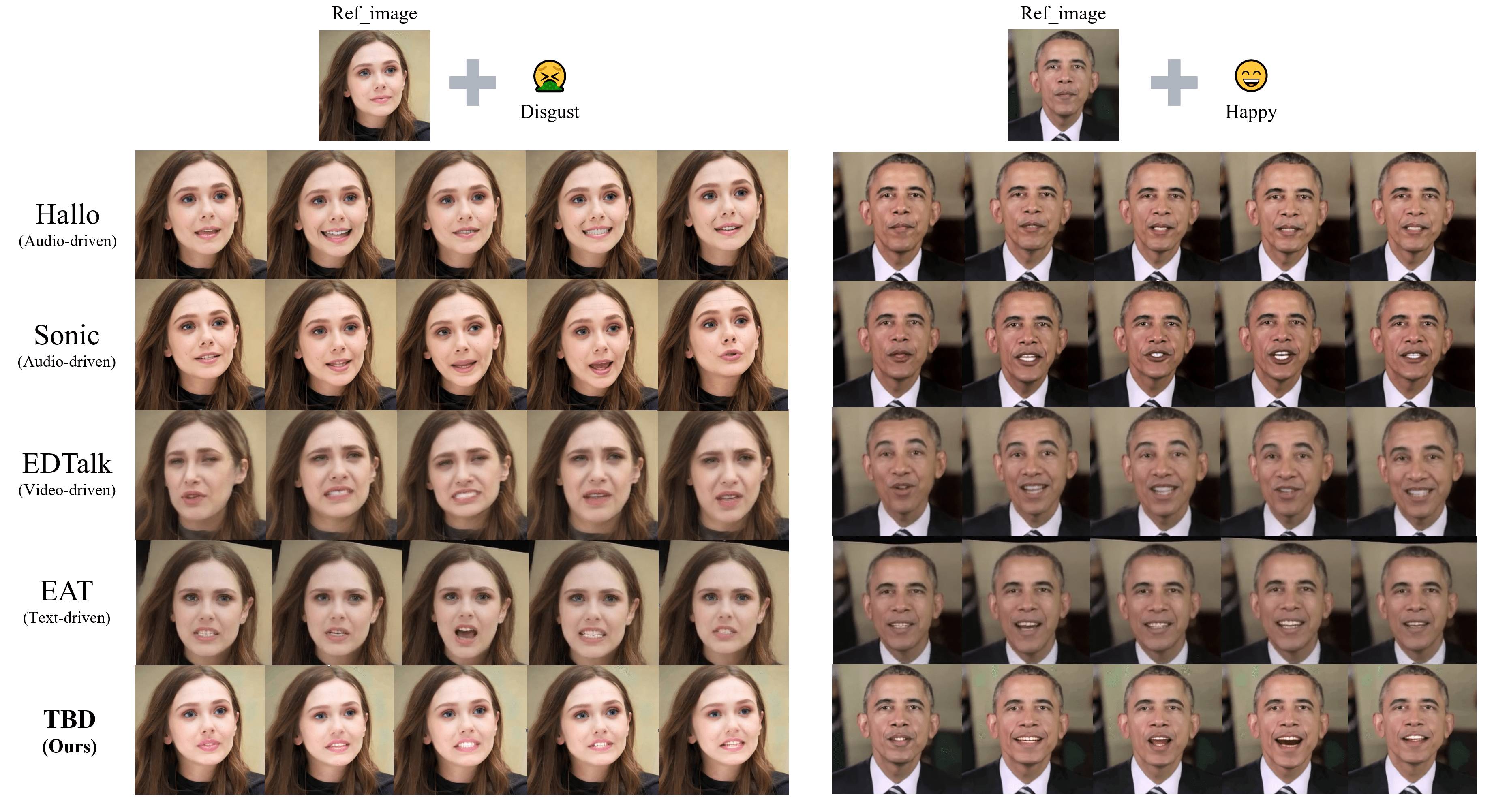}
\caption{Qualitative comparison of facial expression editing based on different inputs. Audio-driven: Hallo, Sonic. Video-driven: EDTalk. Text-driven: EAT, TBD. The results show that our method achieves well emotion controllability with text prompt.}
\label{fig_7}
\end{figure}

\subsection{Qualitative results}
We constructed an open-set test dataset containing over ten portrait images for qualitative evaluation of the proposed method. In comparative experiments, we selected three typical driving approaches as baseline methods: audio-driven, video-driven, text-driven. In the audio-driven approach, we used audio segments with different emotional types as driving conditions. The video-driven method employed video clips with varying emotional tones as input. The text-driven method utilized audio and emotional lable-text as the driving signals.

The experimental results demonstrate that (\autoref{fig_7}): the facial expressions generated by audio-driven methods (Hallo~\cite{hallo} and Sonic~\cite{sonic}) exhibited low alignment with the target emotions, failing to accurately reflect the affective features in speech. 
The EDTalk~\cite{Edtalk} method, driven by reference videos, achieved richer dynamic expressions but suffered from identity inconsistency, causing mismatches between the generated expressions and the target subject's characteristics.
The EAT~\cite{gan2023EAT} method could produce predefined emotional expressions, but due to its reliance on fixed head pose calibration, the generated results deviated from the reference image's pose, with insufficient temporal coherence and naturalness. 
In contrast, our proposed method maintains identity consistency while enabling more natural and precise facial expression editing, demonstrating superior visual expressiveness.

\subsection{Ablation study}
\subsubsection{Efftctiveness of CoT-FA and progressive guidance denoising strategy.}
\begin{figure}[!t]
\centering
\includegraphics[width=5.5 in]{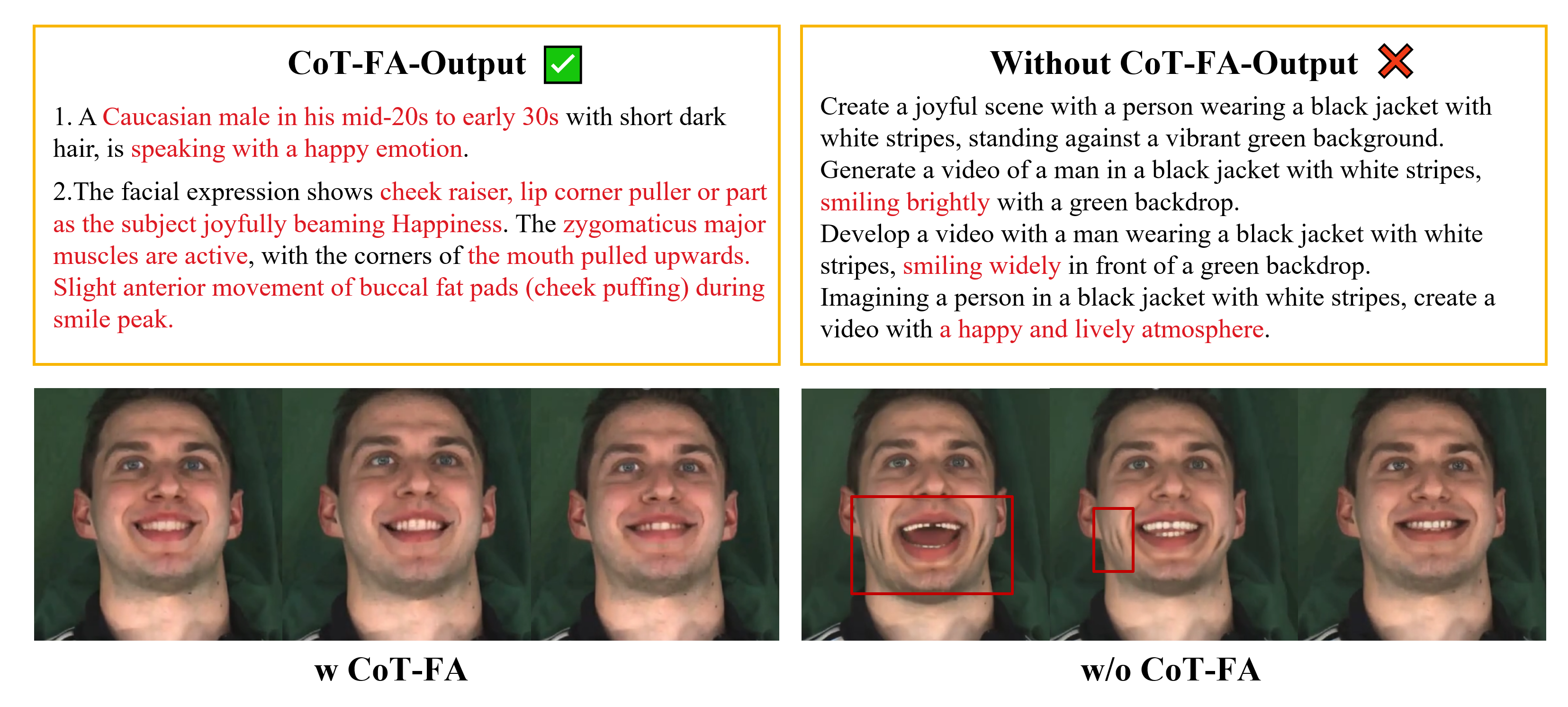}
\captionsetup{font=normalsize}
\caption{Visualization of prompts with CoT-FA and without CoT-FA}
\label{fig_8}
\end{figure}

We conducted systematic ablation expreiments to evaluate the impact of our method. We first validated the effectiveness of the CoT-FA module. As shown in \autoref{fig_8}, prompt outputs generated without this module contained redundant information and exhibited coarser descriptions, whereas those with CoT-FA demonstrated significantly improved precision and granularity. Such differences were also visually observable in the generated videos. Quantitative metrics further confirmed consistent performance gains across multiple indicators when the CoT-FA module was incorporated.

As shown in \autoref{tab_2}, 
the baseline model employs text derived from emotion labels, while ``describ" refers to descriptions obtained through direct question-answering inference (without CoT-FA) using Qwen2-VL on the input image and emotion labels.
The full model (integrating both chain-of-thought facial animation and progressive guidance denoising strategy) demonstrates significant advantages across multiple objective evaluation metrics. These improvements are specifically reflected in key indicators including: FID, FVD~\cite{fvd}, LPIPS~\cite{LPIPS}, SynC. These results validate the synergistic effect of our proposed dual mechanisms in enhancing generation performance.

\begin{table}[ht]
\centering
\captionsetup{font=normalsize}
\caption{The efftctiveness of CoT-FA and progressive guidance denoising strategy.}
\begin{tabular}{l|cccc}
\toprule
Models & FID$\downarrow$ & FVD$\downarrow$ & LPIPS$\downarrow$ & Syn$_{conf}\uparrow$ \\
\midrule
baseline & 20.101 & 431.453 & 0.184 & 4.010 \\
+ describ & 19.159 & 407.589 & 0.180 & 4.090 \\
+ CoT-FA & 18.752 & 399.458 & 0.177 & 4.193 \\
+ CoT-FA + Pgd strategy (full model) & \textbf{16.725} & \textbf{379.16} & \textbf{0.176} & \textbf{4.408} \\
\bottomrule
\end{tabular}
\label{tab_2}
\end{table}

\subsubsection{Analysis of progressive guidance denoising strategy.}
\begin{figure}[!t]
\centering
\includegraphics[width=5 in]{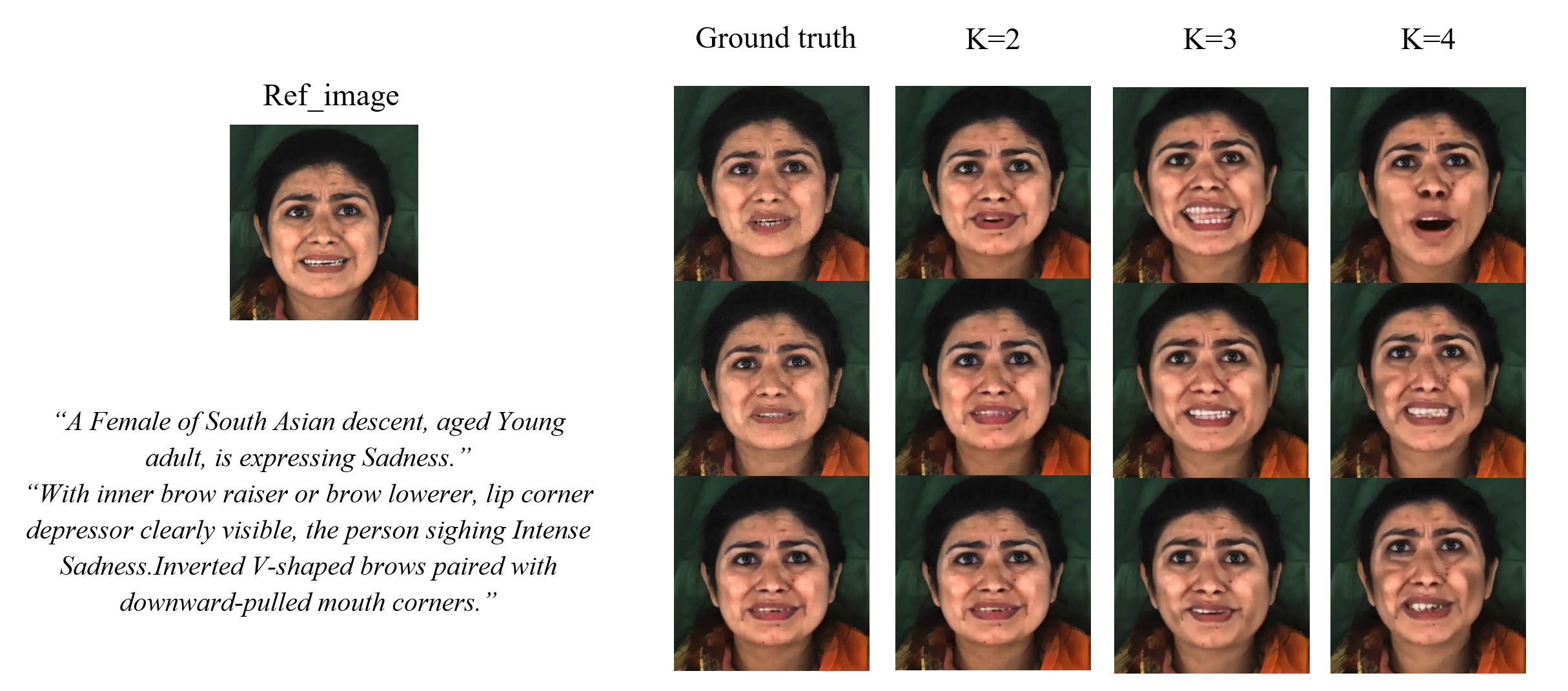}
\captionsetup{font=normalsize}
\caption{Visualization of different progressive guidance denoising strategy. Results demonstrate that the best is achieved when k=2, where two different text prompts guide the generation at denoising steps.}
\label{fig_9}
\end{figure}

We conducted an in-depth analysis of the proposed progressive denoising strategy, with a particular focus on optimizing the partitioning scheme of guiding texts. As shown in \autoref{fig_9}. Experimental comparisons were performed among two-stage, three-stage, and four-stage partitioning approaches, with results demonstrating that the two-stage partitioning yields optimal generation quality. This phenomenon can be explained as follows: given a fixed total denoising step count of 40, excessive stage partitioning would lead to frequent switching of guiding texts, thereby compromising generation stability. In contrast, the two-stage partitioning effectively captures both holistic expression characteristics while preserving sufficient steps for learning detailed features. 

\begin{figure}[!t]
\centering
\includegraphics[width=4.5 in]{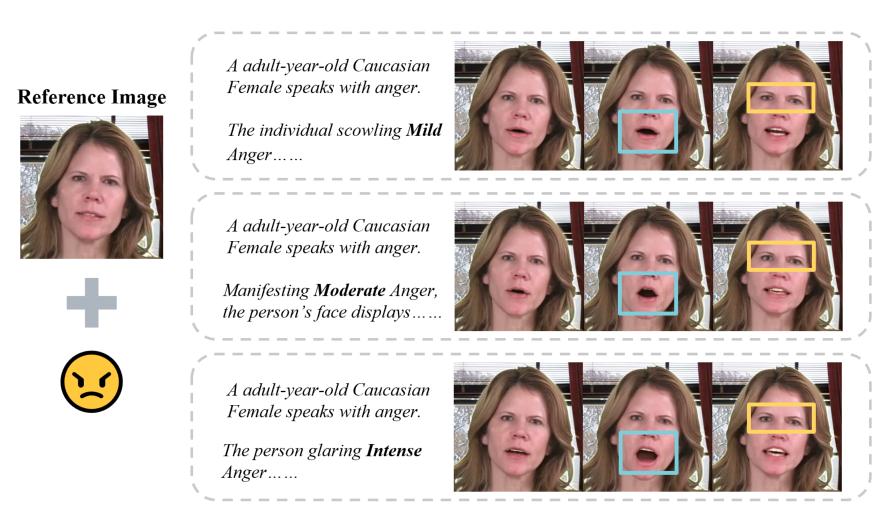}
\captionsetup{font=normalsize}
\caption{Visualization of different intensity control. The three intensity levels of textual descriptions—``Mild", ``Moderate", and ``Intense"—are used to guide the generation. The results show that distinct anger expressions corresponding to each intensity level from the mouth and eyebrow regions.}
\label{fig_10}
\end{figure}

\subsubsection{Intensity control}

We constructed a fine-grained facial expression semantic decomposition framework based on the multi-level intensity annotations (three intensity levels: 1, 2, 3) provided by the MEAD dataset, with each level respectively described as ``Mild", ``Moderate", and ``Intense". As illustrated in \autoref{fig_10}, this graded description mechanism is systematically integrated into the CoT decomposition process, enabling the model to precisely comprehend and generate facial expressions with varying intensity levels. Experimental results demonstrate that our method successfully achieves fine-grained expression control through intensity-based textual descriptions, providing a more precise semantic modulation dimension for facial expression generation tasks.

\subsection{Style code visualization}
We employ t-SNE~\cite{t-sne} to project style codes from the MEAD dataset into a 2D space for visualization. To assess identity consistency, we randomly selected 8 speakers, each with 15 randomly sampled generated videos. For identity feature extraction, we utilized the insightface~\cite{arcface} extractor. As illustrated in \autoref{fig:a}, the identity features of each speaker exhibit strong consistency. For facial expression visualization, we analyzed seven distinct expressions (excluding neutral) from the same speaker, with 20 randomly selected videos per expression. Facial expression features were extracted using InceptionResnet~\cite{inception}. \autoref{fig:b} indicates that while fear and surprise—due to their similar facial muscle movements—show minor overlap, all other expressions are well-separated in the feature space.
\begin{figure*}[htbp]
\captionsetup{font=normalsize}
    \centering
    \begin{minipage}{0.48\textwidth}
        \centering
        \includegraphics[scale=0.6]{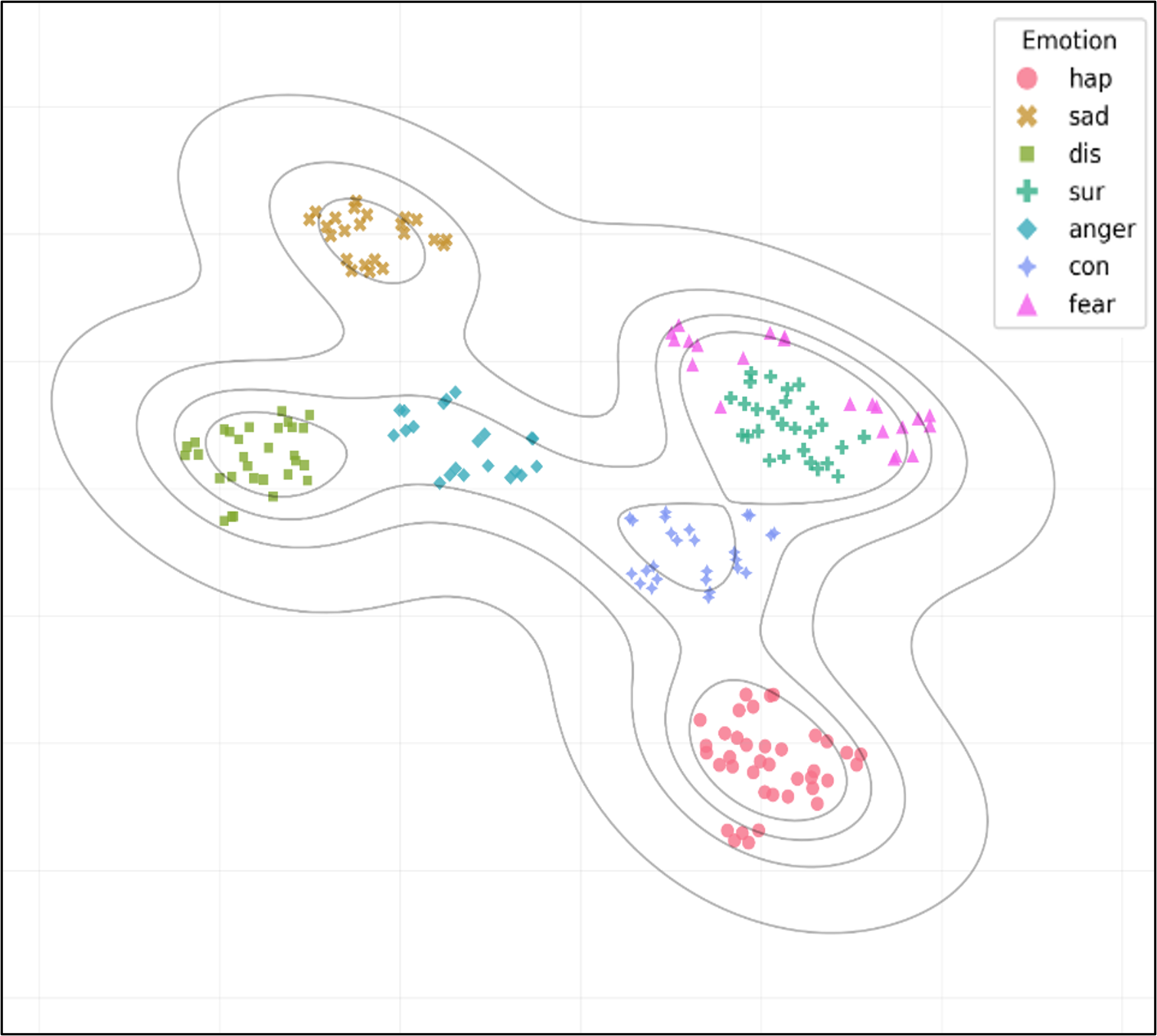}
        \subcaption{Style codes from a speaker} 
        \label{fig:a}
    \end{minipage}
    \hfill
    \begin{minipage}{0.48\textwidth}
        \centering
        \includegraphics[scale=0.6]{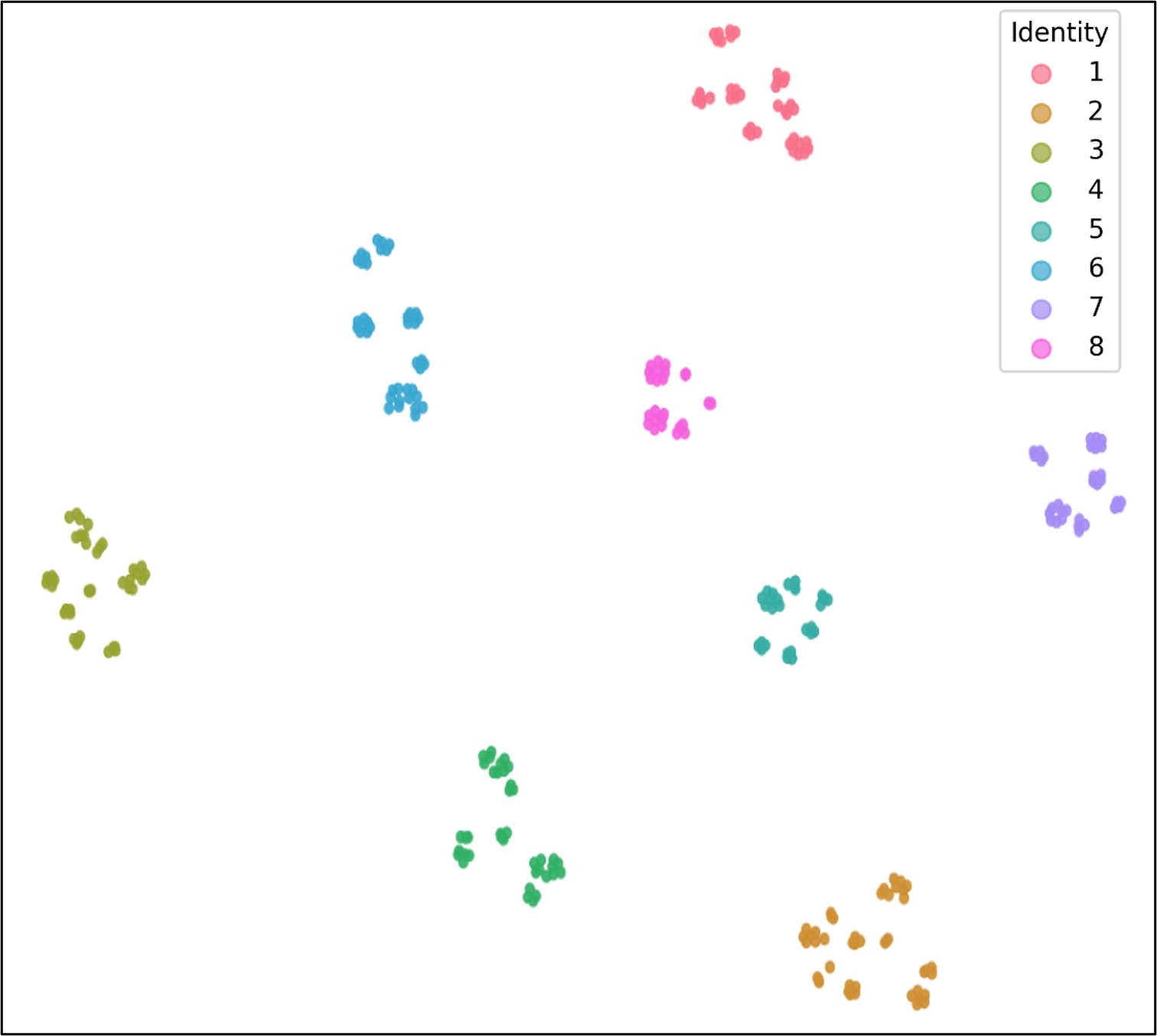}
        \subcaption{Style codes from 8 speakers}
        \label{fig:b}
    \end{minipage}
    \caption{t-SNE visualization of style codes. (a) Emotion-related clustering: Style codes from a single speaker, colored by emotion categories (hap: happy, sad: sadness, dis: disgust, sur: surprise, anger: anger, con: contempt, fear: fear). (b) Identity-related dispersion: Style codes from 8 speakers, colored by speaker identity (ID 1–8).}
    \label{fig:latent_space}
\end{figure*}

\subsection{User study}
To comprehensively evaluate the performance of our method, we conducted a user study comparing it with other state-of-the-art emotion-aware talking head generation approaches. The study involved 10 experienced evaluators who assessed both generation quality and controllability. For evaluation, we randomly selected 10 generation samples from both the MEAD dataset and an open-set dataset. Our assessment employed the Mean Opinion Score (MOS) metric across four key dimensions: (1) Lip-sync accuracy (Lip.) - measuring the synchronization between speech and mouth movements. (2) Emotion controllability (Emo.) - evaluating the consistency between generated facial expressions and target emotions. (3) Naturalness (Nat.) - assessing whether the talking head maintains realistic and fluid facial dynamics during emotional speech. (4) Identity preservation (ID.) - verifying the faithfulness to the reference identity's speaking characteristics. 

During the evaluation, participants were presented with one video at a time and asked to rate each video on a 5-point Likert scale (1 being poorest, 5 being excellent) for all four criteria. The final scores were calculated by averaging all ratings. 

As demonstrated in \autoref{tab_3}, our proposed method outperforms all baseline approaches across every evaluation metric, achieving superior results in lip synchronization, emotion control, motion naturalness, and identity preservation. This comprehensive evaluation validates the effectiveness of our approach in generating high-quality, emotion-controllable talking head videos.

\begin{table}[h]
\centering
\captionsetup{font=normalsize}
\caption{User Study for TBD and other baselines on MEAD and open-set dataset}
\label{tab:user_study}
\setlength{\tabcolsep}{12pt} 
\renewcommand{\arraystretch}{1.2} 
\begin{tabular}{lcccc}
\toprule
\textbf{Method} & \textbf{Lip.$\uparrow$} & \textbf{Nat.$\uparrow$} & \textbf{Emo.$\uparrow$} & \textbf{ID.$\uparrow$} \\
\midrule
EDTalk & 3.87 & 3.37 & 3.31 & 3.04 \\
ETA & 3.87 & 3.43 & 3.12 & 3.77 \\
\textbf{TBD (Ours)} & \textbf{4.00} & \textbf{3.58} & \textbf{3.40} & \textbf{3.86} \\
\bottomrule
\end{tabular}
\label{tab_3}
\end{table}

\section{Limitations}
While Our study has made several advances in emotion-aware talking head generation, certain limitations remain that warrant further exploration and consideration. These limitations highlight potential directions for future research to refine and extend the proposed method: (1) Enhancing emotional expressiveness naturalness. Developing more comprehensive affective expression systems by incorporating nonverbal cues such as head poses and eye movements, while rigorously maintaining identity consistency, could significantly improve the realism and expressiveness of generated results. (2) Improving audio-visual synchronization. Current methods still exhibit insufficient extraction and mapping of speech emotional features. Future work should focus on deeper utilization of speech information to enhance facial expression generation, establishing more precise audio-visual correspondence to improve both lip-sync accuracy and emotional expression alignment. (3) Advancing model architecture. The UNet-based diffusion model has inherent computational efficiency limitations. Promising research directions include adaptive improvements of novel architectures like Diffusion Transformers (DiT)~\cite{DiT}, leveraging their global attention mechanisms and scalability advantages to significantly boost inference speed while maintaining generation quality.

\section{Conclusion}
This paper presents Think-Before-Draw, a novel framework for emotion-aware talking-head generation that leverages Chain-of-Thought and a progressive guidance denoising strategy to achieve fine-grained, naturalistic facial animation under text guidance. The proposed multi-step semantic decomposition transforms discrete emotion labels into FACS-aligned high-level semantics, enabling more and nuanced emotional generation. By introducing a progressive guidance denoising strategy—first establishing an emotional ground state and then refining localized muscle dynamics—our method mimics the artistic process of ``global composition, local refinement" for enhanced realism.
Our experiments demonstrate that our approach achieves state-of-the-art performance on widely-used benchmarks, including MEAD
and HDTF. Additionally, we collected a set of portrait images to evaluate our model’s zero-shot generation capability.
Extensive quantitative and qualitative evaluations demonstrate that Think-Before-Draw outperforms existing methods in emotional expressiveness, motion naturalness, and user controllability. This work provides a theoretically grounded and practically effective solution for fine-grained emotional control in virtual humans, contributing to more natural and expressive human-computer interaction. Future research may explore adaptive emotion modeling for personalized avatars and real-time synthesis optimizations to further advance the field.










\clearpage
\appendix
\section{The Networks Details}
\setcounter{figure}{0}
Our training model is based on Stable Diffusion, where we employ ReferenceNet to extract features from reference images, as illustrated in \autoref{fig:a} \autoref{fig:b}. Similar to the denoising UNet, ReferenceNet inherits its initial weights from the original Stable Diffusion, with independent weight updates performed for each module. We then detail the method for integrating features from ReferenceNet into the denoising UNet. Specifically, the self-attention layer is replaced with a reference-attention layer to enable cross-modality feature interaction. Given a feature map 
$x_d\in\mathbb{R}^{t\times{h}\times{w}\times{c}}$ from denoising UNet and $x_r\in\mathbb{R}^{h\times{w}\times{c}}$ from Reference UNet, we first copy $x_r$ by $t$ times and concatenate it with $x_d$ along $w$ dimension. 

\begin{figure}[H]  
    \centering
    
    \begin{subfigure}[b]{0.45\textwidth}
        \centering
        \includegraphics[width=\textwidth]{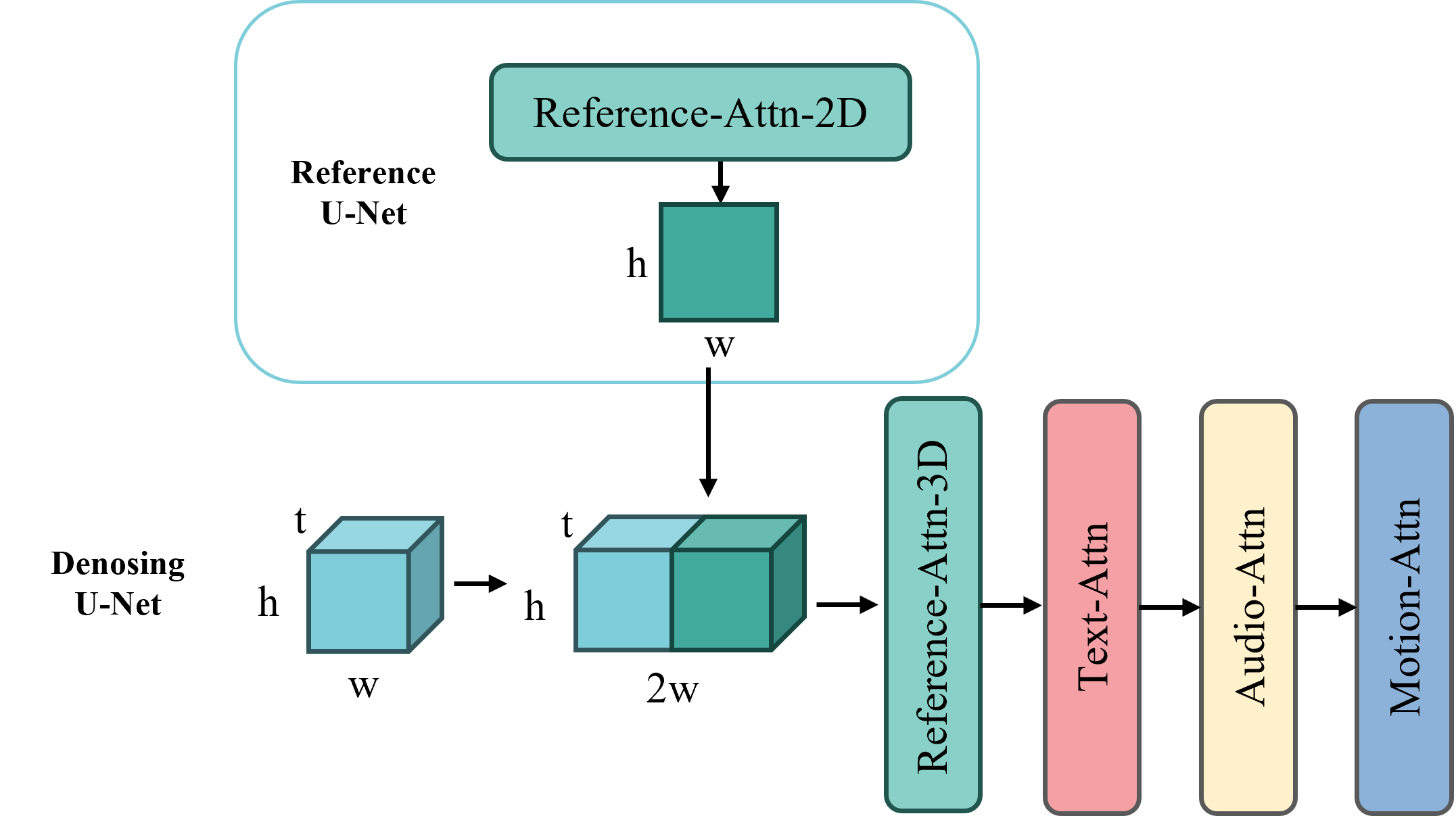}
        \caption{CrossAttnBlock}
        \label{fig:a}
    \end{subfigure}
    \hfill
    \begin{subfigure}[b]{0.35\textwidth}
        \centering
        \includegraphics[width=\textwidth]{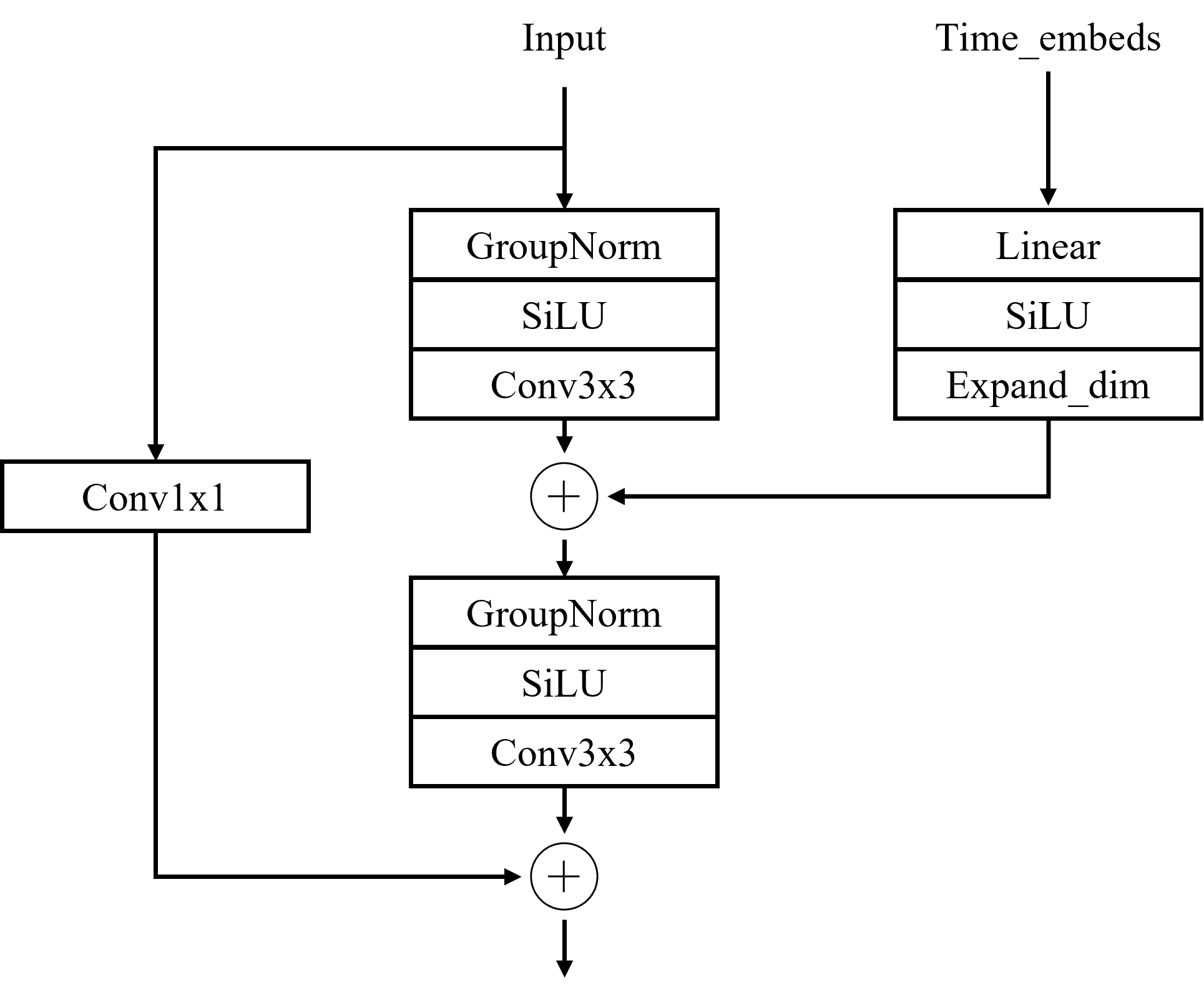}
        \caption{Reference-Attention Layer}
        \label{fig:b}
    \end{subfigure}
    
    \vspace{0.5cm}  
    
    \begin{subfigure}[b]{0.3\textwidth}
        \centering
        \includegraphics[width=\textwidth]{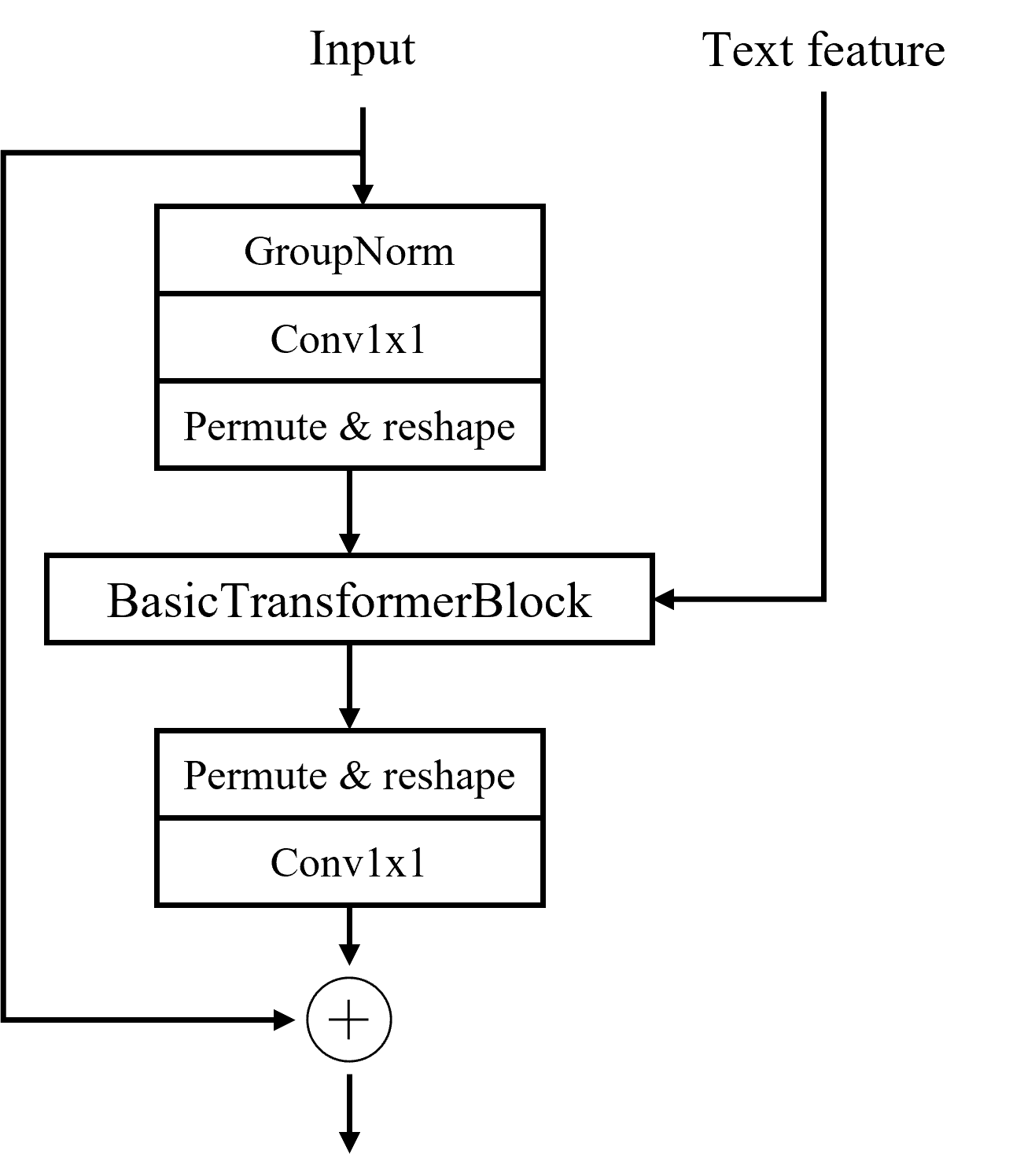}
        \caption{Text-Attention Layer}
        \label{fig:c}
    \end{subfigure}
    \hfill
    \begin{subfigure}[b]{0.3\textwidth}
        \centering
        \includegraphics[width=\textwidth]{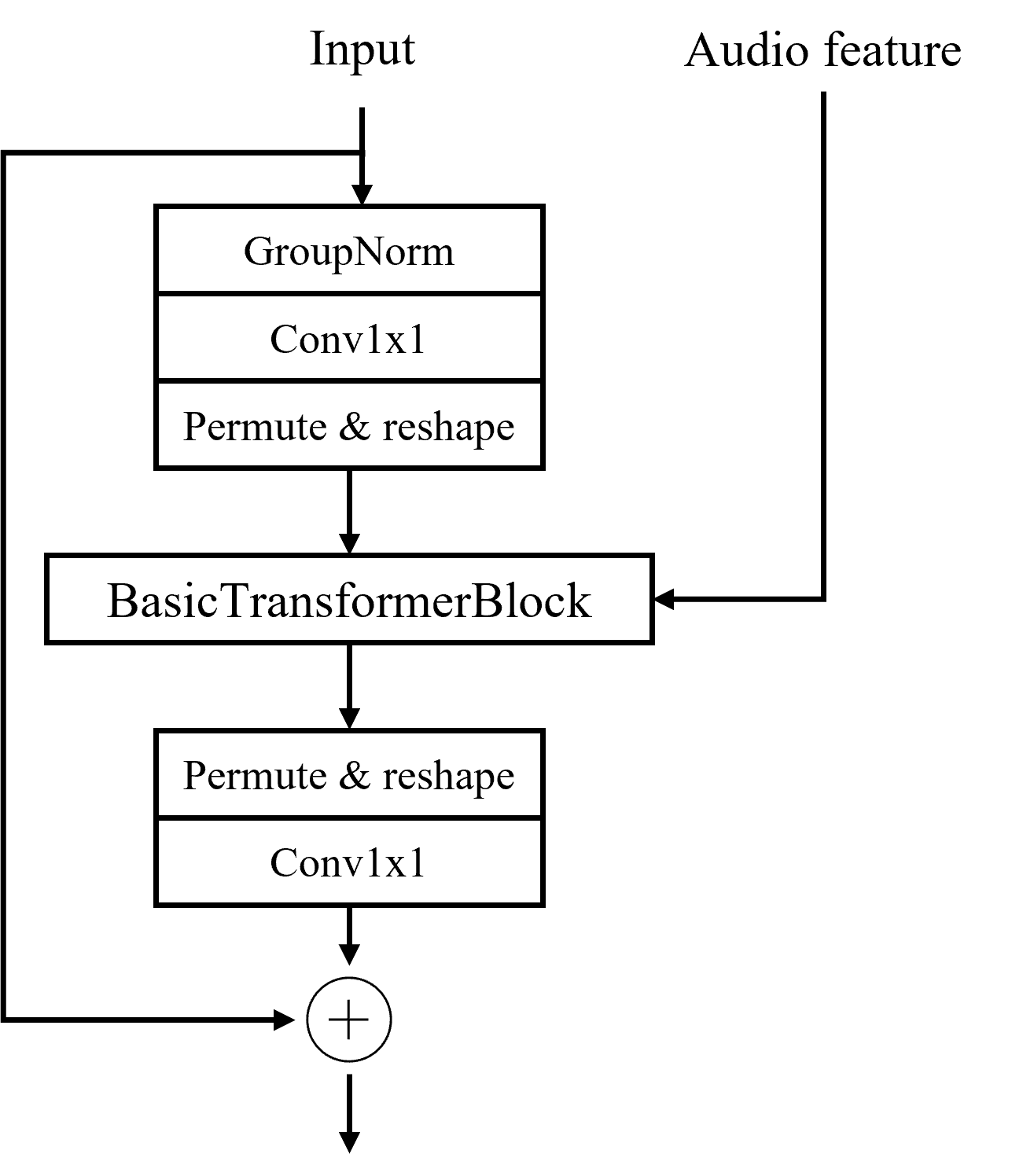}
        \caption{Audio-Attention Layer}
        \label{fig:d}
    \end{subfigure}
    \hfill
    \begin{subfigure}[b]{0.23\textwidth}
        \centering
        \includegraphics[width=\textwidth]{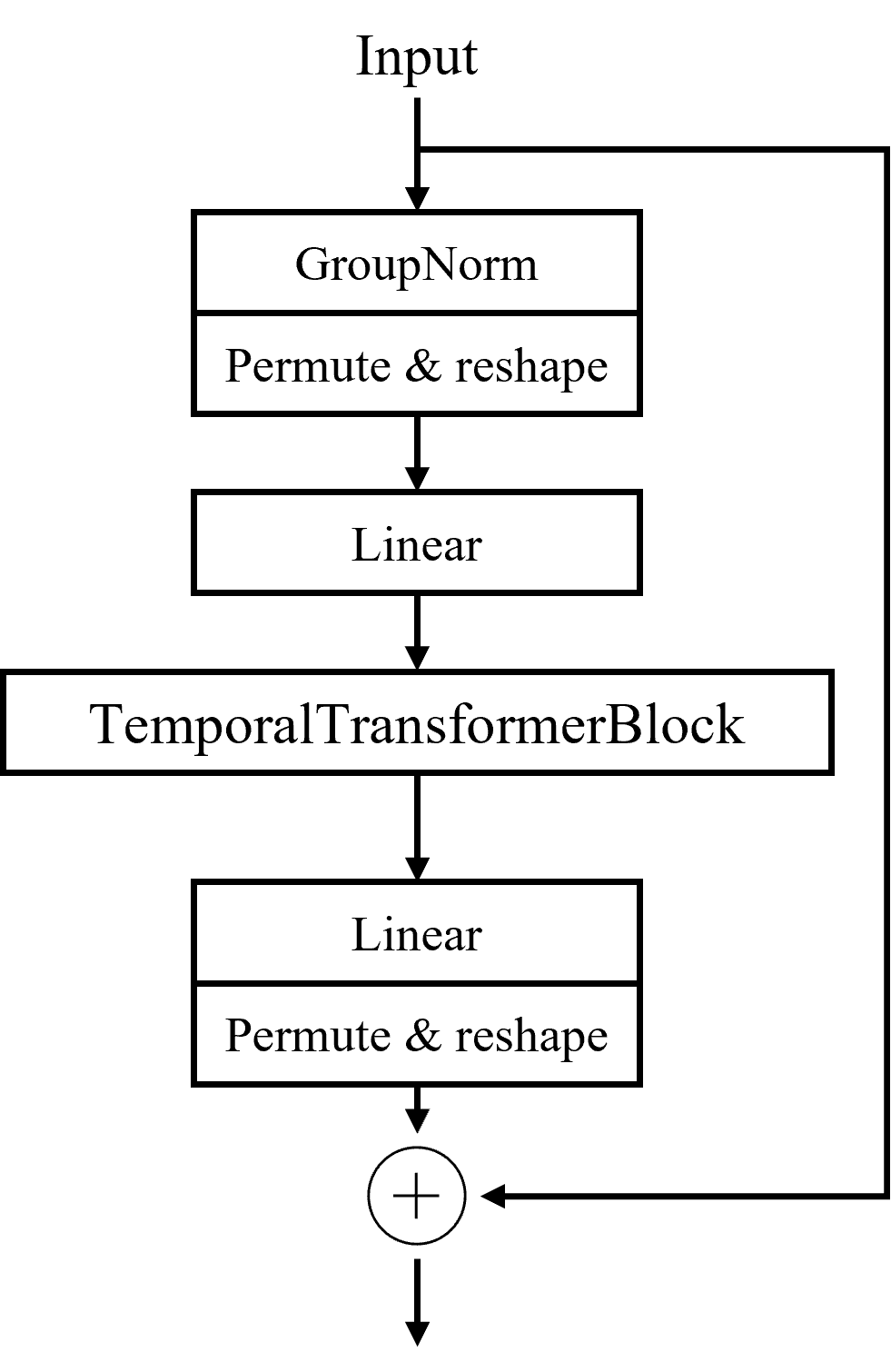}
        \caption{Motion-Attention Layer}
        \label{fig:e}
    \end{subfigure}
    
    \caption{Detailed architecture in our TBD}
    \label{fig:all}
\end{figure}

\textbf{Text-Attention Layer} As shown in \autoref{fig:c}, the input to the Text-Attention Layer is derived from the output of the Reference-Attention Layer. It first passes through a GroupNorm layer to normalize the data and stabilize the training process. Next, it undergoes a Conv1x1 layer, which adjusts the channel dimensions or applies a linear transformation. The data then proceeds through a Permute \& Reshape operation to reorganize its dimensional order and shape for subsequent modules. Subsequently, it enters a BasicTransformerBlock to capture both long- and short-term dependencies. After another Permute \& Reshape operation and Conv1x1 layer, a residual connection (via addition) is applied with the original input to produce the final output.

\textbf{Audio-Attention Layer} As illustrated in \autoref{fig:d}, the Audio-Attention Layer takes the output of the Text-Attention Layer as its input. Its structure is identical to the Text-Attention Layer, except that cross-attention is conditioned on audio features.

\textbf{Motion-Attention Layer} As depicted in \autoref{fig:e}, the Motion-Attention Layer receives input from the Audio-Attention Layer. The data first undergoes normalization via the GroupNorm layer, followed by a Permute \& Reshape operation. It then passes through a Linear layer for linear transformation before entering the TemporalTransformerBlock to process temporal dependencies. Another Linear layer and Permute \& Reshape operation follow, and finally, a residual connection (via addition) merges the result with the original input to generate the output.

\section{Additional Experimental Results}
\setcounter{figure}{0}
\begin{figure}[h]
\centering
\includegraphics[width=6 in]{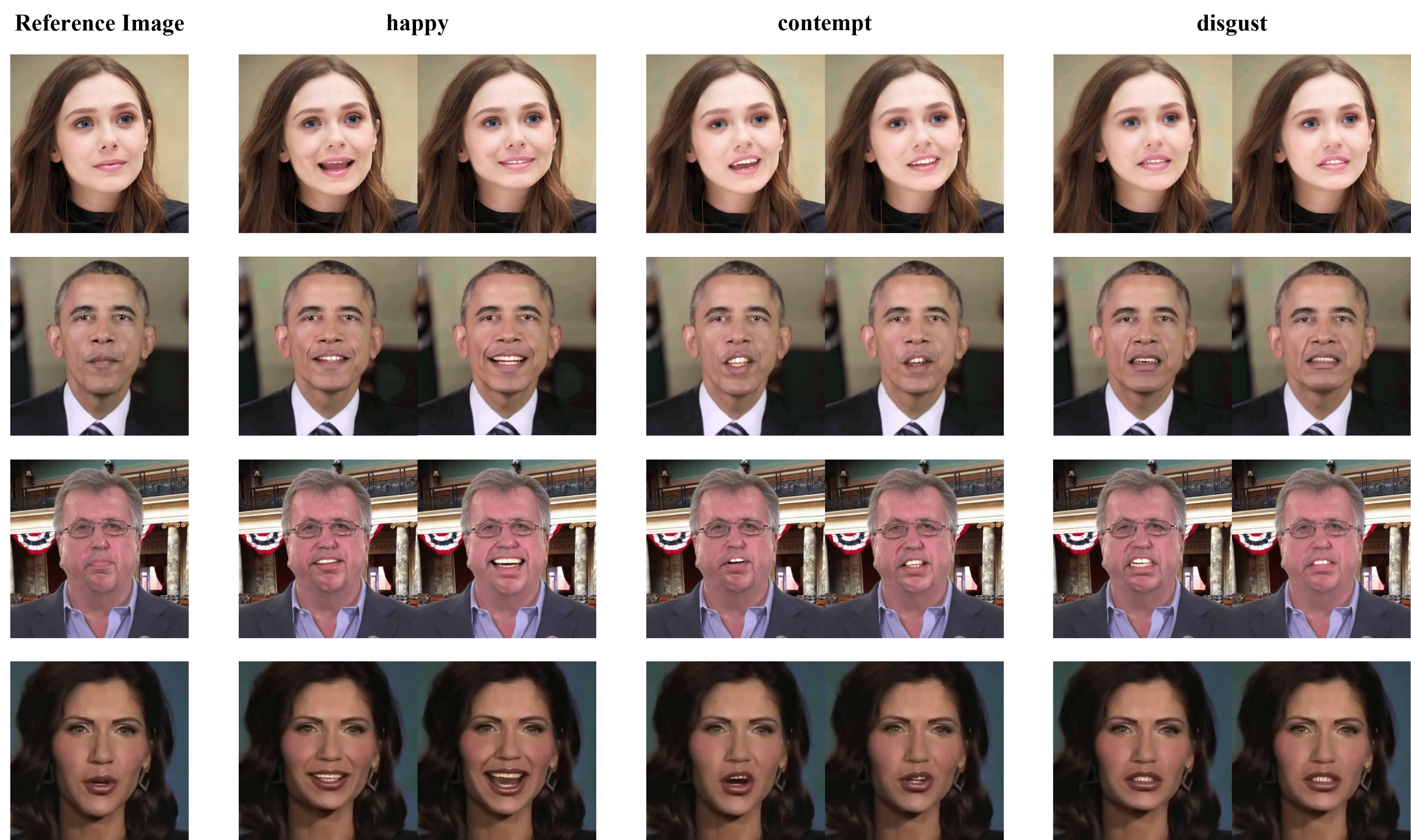}
\captionsetup{font=normalsize}
\caption{More zero-shot generation results}
\label{fig_13}
\end{figure}
\begin{figure}[h]
\centering
\includegraphics[width=6 in]{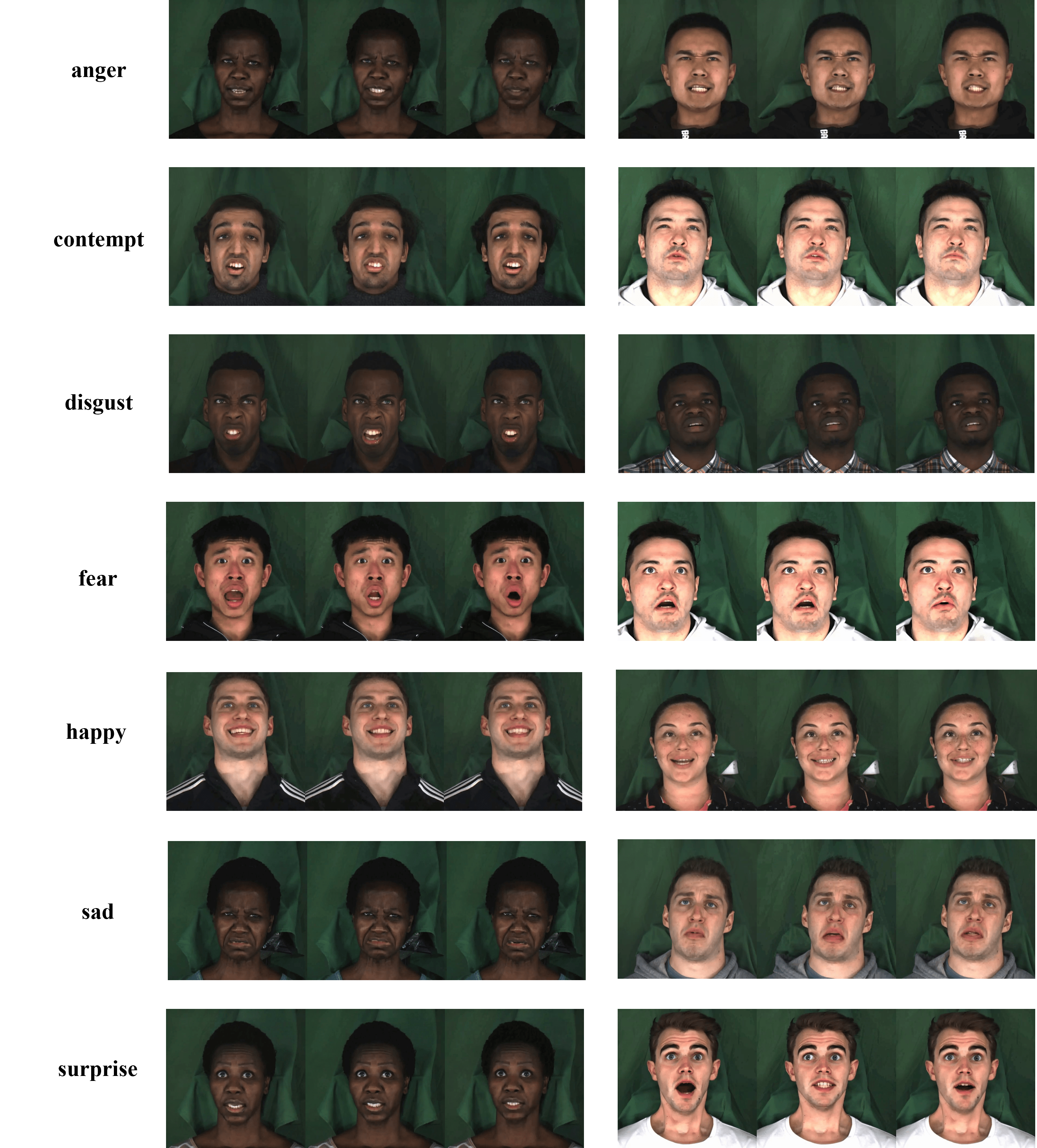}
\captionsetup{font=normalsize}
\caption{More Qualitative results}
\label{fig_14}
\end{figure}

\clearpage

\bibliographystyle{elsarticle-num} 
\bibliography{mylib.bib}

\begin{thebibliography}{10}
\expandafter\ifx\csname url\endcsname\relax
  \def\url#1{\texttt{#1}}\fi
\expandafter\ifx\csname urlprefix\endcsname\relax\def\urlprefix{URL }\fi
\expandafter\ifx\csname href\endcsname\relax
  \def\href#1#2{#2} \def\path#1{#1}\fi

\bibitem{metaverse}
S.~Mystakidis, Metaverse, Encyclopedia 2~(1) (2022) 486--497.

\bibitem{metaversesurvey}
H.~Wang, H.~Ning, Y.~Lin, W.~Wang, S.~Dhelim, F.~Farha, J.~Ding, M.~Daneshmand, A survey on the metaverse: The state-of-the-art, technologies, applications, and challenges, IEEE Internet of Things Journal 10~(16) (2023) 14671--14688.

\bibitem{zhen2023human}
R.~Zhen, W.~Song, Q.~He, J.~Cao, L.~Shi, J.~Luo, Human-computer interaction system: A survey of talking-head generation, Electronics 12~(1) (2023) 218.

\bibitem{zhou2024survey}
P.~Zhou, L.~Wang, Z.~Liu, Y.~Hao, P.~Hui, S.~Tarkoma, J.~Kangasharju, A survey on generative ai and llm for video generation, understanding, and streaming, arXiv preprint arXiv:2404.16038 (2024).

\bibitem{ji2021Audio-Driven}
X.~Ji, H.~Zhou, K.~Wang, W.~Wu, C.~C. Loy, X.~Cao, F.~Xu, Audio-driven emotional video portraits, in: Proceedings of the IEEE/CVF conference on computer vision and pattern recognition, 2021, pp. 14080--14089.

\bibitem{ji2022EAMM}
X.~Ji, H.~Zhou, K.~Wang, Q.~Wu, W.~Wu, F.~Xu, X.~Cao, Eamm: One-shot emotional talking face via audio-based emotion-aware motion model, in: ACM SIGGRAPH 2022 conference proceedings, 2022, pp. 1--10.

\bibitem{ma2023styletalk}
Y.~Ma, S.~Wang, Z.~Hu, C.~Fan, T.~Lv, Y.~Ding, Z.~Deng, X.~Yu, Styletalk: One-shot talking head generation with controllable speaking styles, in: Proceedings of the AAAI conference on artificial intelligence, Vol.~37, 2023, pp. 1896--1904.

\bibitem{zhang2023sadtalker}
W.~Zhang, X.~Cun, X.~Wang, Y.~Zhang, X.~Shen, Y.~Guo, Y.~Shan, F.~Wang, Sadtalker: Learning realistic 3d motion coefficients for stylized audio-driven single image talking face animation, in: Proceedings of the IEEE/CVF conference on computer vision and pattern recognition, 2023, pp. 8652--8661.

\bibitem{peng2023emotalk}
Z.~Peng, H.~Wu, Z.~Song, H.~Xu, X.~Zhu, J.~He, H.~Liu, Z.~Fan, Emotalk: Speech-driven emotional disentanglement for 3d face animation, in: Proceedings of the IEEE/CVF international conference on computer vision, 2023, pp. 20687--20697.

\bibitem{gan2023EAT}
Y.~Gan, Z.~Yang, X.~Yue, L.~Sun, Y.~Yang, Efficient emotional adaptation for audio-driven talking-head generation, in: Proceedings of the IEEE/CVF International Conference on Computer Vision, 2023, pp. 22634--22645.

\bibitem{liu2025moee}
H.~Liu, W.~Sun, D.~Di, S.~Sun, J.~Yang, C.~Zou, H.~Bao, Moee: Mixture of emotion experts for audio-driven portrait animation, arXiv preprint arXiv:2501.01808 (2025).

\bibitem{survey-MLLM}
Z.~Liang, Y.~Xu, Y.~Hong, P.~Shang, Q.~Wang, Q.~Fu, K.~Liu, A survey of multimodel large language models, in: Proceedings of the 3rd International Conference on Computer, Artificial Intelligence and Control Engineering, 2024, pp. 405--409.

\bibitem{qwen2}
P.~Wang, S.~Bai, S.~Tan, S.~Wang, Z.~Fan, J.~Bai, K.~Chen, X.~Liu, J.~Wang, W.~Ge, et~al., Qwen2-vl: Enhancing vision-language model's perception of the world at any resolution, arXiv preprint arXiv:2409.12191 (2024).

\bibitem{fg-emotalk}
Z.~Sun, Y.~Xuan, F.~Liu, Y.~Xiang, Fg-emotalk: Talking head video generation with fine-grained controllable facial expressions, in: Proceedings of the AAAI Conference on Artificial Intelligence, Vol.~38, 2024, pp. 5043--5051.

\bibitem{expclip}
Y.~Zhong, H.~Wei, P.~Yang, Z.~Wang, Expclip: Bridging text and facial expressions via semantic alignment, in: Proceedings of the AAAI Conference on Artificial Intelligence, Vol.~38, 2024, pp. 7614--7622.

\bibitem{pan2024expressive}
Y.~Pan, S.~Tan, S.~Cheng, Q.~Lin, Z.~Zeng, K.~Mitchell, Expressive talking avatars, IEEE Transactions on Visualization and Computer Graphics (2024).

\bibitem{ekman1978facial}
P.~Ekman, W.~V. Friesen, Facial action coding system, Environmental Psychology \& Nonverbal Behavior (1978).

\bibitem{wei2022CoT}
J.~Wei, X.~Wang, D.~Schuurmans, M.~Bosma, F.~Xia, E.~Chi, Q.~V. Le, D.~Zhou, et~al., Chain-of-thought prompting elicits reasoning in large language models, Advances in neural information processing systems 35 (2022) 24824--24837.

\bibitem{zhang2022automatic}
Z.~Zhang, A.~Zhang, M.~Li, A.~Smola, Automatic chain of thought prompting in large language models, arXiv preprint arXiv:2210.03493 (2022).

\bibitem{jarlier2011muscle}
S.~Jarlier, D.~Grandjean, S.~Delplanque, K.~N'diaye, I.~Cayeux, M.~I. Velazco, D.~Sander, P.~Vuilleumier, K.~R. Scherer, Thermal analysis of facial muscles contractions, IEEE transactions on affective computing 2~(1) (2011) 2--9.

\bibitem{2005facialanalysis}
S.~Z. Li, A.~K. Jain, Y.-L. Tian, T.~Kanade, J.~F. Cohn, Facial expression analysis, Handbook of face recognition (2005) 247--275.

\bibitem{duchenne1990mechanism}
G.-B. Duchenne, The mechanism of human facial expression, Cambridge university press, 1990.

\bibitem{faigin2012artist}
G.~Faigin, The artist's complete guide to facial expression, Watson-Guptill, 2012.

\bibitem{loomis2021drawing}
A.~Loomis, Drawing the head \& hands, Clube de Autores, 2021.

\bibitem{xu2024hallo}
M.~Xu, H.~Li, Q.~Su, H.~Shang, L.~Zhang, C.~Liu, J.~Wang, Y.~Yao, S.~Zhu, Hallo: Hierarchical audio-driven visual synthesis for portrait image animation, arXiv preprint arXiv:2406.08801 (2024).

\bibitem{echomimic}
Z.~Chen, J.~Cao, Z.~Chen, Y.~Li, C.~Ma, Echomimic: Lifelike audio-driven portrait animations through editable landmark conditions, in: Proceedings of the AAAI Conference on Artificial Intelligence, Vol.~39, 2025, pp. 2403--2410.

\bibitem{sonic}
X.~Ji, X.~Hu, Z.~Xu, J.~Zhu, C.~Lin, Q.~He, J.~Zhang, D.~Luo, Y.~Chen, Q.~Lin, et~al., Sonic: Shifting focus to global audio perception in portrait animation, arXiv preprint arXiv:2411.16331 (2024).

\bibitem{Edtalk}
S.~Tan, B.~Ji, M.~Bi, Y.~Pan, Edtalk: Efficient disentanglement for emotional talking head synthesis, in: European Conference on Computer Vision, Springer, 2024, pp. 398--416.

\bibitem{wei2022emergent}
J.~Wei, Y.~Tay, R.~Bommasani, C.~Raffel, B.~Zoph, S.~Borgeaud, D.~Yogatama, M.~Bosma, D.~Zhou, D.~Metzler, et~al., Emergent abilities of large language models, arXiv preprint arXiv:2206.07682 (2022).

\bibitem{prystawski2023think}
B.~Prystawski, M.~Li, N.~Goodman, Why think step by step? reasoning emerges from the locality of experience, Advances in Neural Information Processing Systems 36 (2023) 70926--70947.

\bibitem{MM-COT}
Z.~Zhang, A.~Zhang, M.~Li, H.~Zhao, G.~Karypis, A.~Smola, Multimodal chain-of-thought reasoning in language models, arXiv preprint arXiv:2302.00923 (2023).

\bibitem{visual-cot}
H.~Shao, S.~Qian, H.~Xiao, G.~Song, Z.~Zong, L.~Wang, Y.~Liu, H.~Li, Visual cot: Advancing multi-modal language models with a comprehensive dataset and benchmark for chain-of-thought reasoning, Advances in Neural Information Processing Systems 37 (2024) 8612--8642.

\bibitem{xie2024show-O}
J.~Xie, W.~Mao, Z.~Bai, D.~J. Zhang, W.~Wang, K.~Q. Lin, Y.~Gu, Z.~Chen, Z.~Yang, M.~Z. Shou, Show-o: One single transformer to unify multimodal understanding and generation, arXiv preprint arXiv:2408.12528 (2024).

\bibitem{guo2025GoT}
Z.~Guo, R.~Zhang, C.~Tong, Z.~Zhao, P.~Gao, H.~Li, P.-A. Heng, Can we generate images with cot? let's verify and reinforce image generation step by step, arXiv preprint arXiv:2501.13926 (2025).

\bibitem{fei2024dysen}
H.~Fei, S.~Wu, W.~Ji, H.~Zhang, T.-S. Chua, Dysen-vdm: Empowering dynamics-aware text-to-video diffusion with llms, in: Proceedings of the IEEE/CVF Conference on Computer Vision and Pattern Recognition, 2024, pp. 7641--7653.

\bibitem{dsg}
J.~Ji, R.~Krishna, L.~Fei-Fei, J.~C. Niebles, Action genome: Actions as compositions of spatio-temporal scene graphs, in: Proceedings of the IEEE/CVF conference on computer vision and pattern recognition, 2020, pp. 10236--10247.

\bibitem{wav2vec}
S.~Schneider, A.~Baevski, R.~Collobert, M.~Auli, wav2vec: Unsupervised pre-training for speech recognition, arXiv preprint arXiv:1904.05862 (2019).

\bibitem{VAE}
D.~P. Kingma, M.~Welling, et~al., Auto-encoding variational bayes (2013).

\bibitem{clip}
A.~Radford, J.~W. Kim, C.~Hallacy, A.~Ramesh, G.~Goh, S.~Agarwal, G.~Sastry, A.~Askell, P.~Mishkin, J.~Clark, et~al., Learning transferable visual models from natural language supervision, in: International conference on machine learning, PmLR, 2021, pp. 8748--8763.

\bibitem{hallo}
M.~Xu, H.~Li, Q.~Su, H.~Shang, L.~Zhang, C.~Liu, J.~Wang, Y.~Yao, S.~Zhu, Hallo: Hierarchical audio-driven visual synthesis for portrait image animation, arXiv preprint arXiv:2406.08801 (2024).

\bibitem{stablediffusion}
A.~Blattmann, T.~Dockhorn, S.~Kulal, D.~Mendelevitch, M.~Kilian, D.~Lorenz, Y.~Levi, Z.~English, V.~Voleti, A.~Letts, et~al., Stable video diffusion: Scaling latent video diffusion models to large datasets, arXiv preprint arXiv:2311.15127 (2023).

\bibitem{Animatediff}
Y.~Guo, C.~Yang, A.~Rao, Z.~Liang, Y.~Wang, Y.~Qiao, M.~Agrawala, D.~Lin, B.~Dai, Animatediff: Animate your personalized text-to-image diffusion models without specific tuning, arXiv preprint arXiv:2307.04725 (2023).

\bibitem{DDPM}
J.~Ho, A.~Jain, P.~Abbeel, Denoising diffusion probabilistic models, Advances in neural information processing systems 33 (2020) 6840--6851.

\bibitem{DDIM}
J.~Song, C.~Meng, S.~Ermon, Denoising diffusion implicit models, arXiv preprint arXiv:2010.02502 (2020).

\bibitem{wang2020mead}
K.~Wang, Q.~Wu, L.~Song, Z.~Yang, W.~Wu, C.~Qian, R.~He, Y.~Qiao, C.~C. Loy, Mead: A large-scale audio-visual dataset for emotional talking-face generation, in: European conference on computer vision, Springer, 2020, pp. 700--717.

\bibitem{HDTF}
Z.~Zhang, L.~Li, Y.~Ding, C.~Fan, Flow-guided one-shot talking face generation with a high-resolution audio-visual dataset, in: Proceedings of the IEEE/CVF conference on computer vision and pattern recognition, 2021, pp. 3661--3670.

\bibitem{makelttalk}
Y.~Zhou, X.~Han, E.~Shechtman, J.~Echevarria, E.~Kalogerakis, D.~Li, Makelttalk: speaker-aware talking-head animation, ACM Transactions On Graphics (TOG) 39~(6) (2020) 1--15.

\bibitem{PCAVS}
H.~Zhou, Y.~Sun, W.~Wu, C.~C. Loy, X.~Wang, Z.~Liu, Pose-controllable talking face generation by implicitly modularized audio-visual representation, in: Proceedings of the IEEE/CVF conference on computer vision and pattern recognition, 2021, pp. 4176--4186.

\bibitem{PSNR}
Q.~Huynh-Thu, M.~Ghanbari, Scope of validity of psnr in image/video quality assessment, Electronics letters 44~(13) (2008) 800--801.

\bibitem{SSIM}
Z.~Wang, A.~C. Bovik, H.~R. Sheikh, E.~P. Simoncelli, Image quality assessment: from error visibility to structural similarity, IEEE transactions on image processing 13~(4) (2004) 600--612.

\bibitem{Seitzer2020FID}
M.~Seitzer, {pytorch-fid: FID Score for PyTorch}, \url{https://github.com/mseitzer/pytorch-fid}, version 0.3.0 (August 2020).

\bibitem{CPBD_Alt}
N.~D. Narvekar, L.~J. Karam, A no-reference image blur metric based on the cumulative probability of blur detection (cpbd), IEEE Transactions on Image Processing 20~(9) (2011) 2678--2683.
\newblock \href {https://doi.org/10.1109/TIP.2011.2131660} {\path{doi:10.1109/TIP.2011.2131660}}.

\bibitem{sync}
J.~S. Chung, A.~Zisserman, Out of time: automated lip sync in the wild, in: Computer Vision--ACCV 2016 Workshops: ACCV 2016 International Workshops, Taipei, Taiwan, November 20-24, 2016, Revised Selected Papers, Part II 13, Springer, 2017, pp. 251--263.

\bibitem{wav2lip-data}
T.~Afouras, J.~S. Chung, A.~Senior, O.~Vinyals, A.~Zisserman, Deep audio-visual speech recognition, IEEE transactions on pattern analysis and machine intelligence 44~(12) (2018) 8717--8727.

\bibitem{fvd}
T.~Unterthiner, S.~Van~Steenkiste, K.~Kurach, R.~Marinier, M.~Michalski, S.~Gelly, Fvd: A new metric for video generation (2019).

\bibitem{LPIPS}
R.~Zhang, P.~Isola, A.~A. Efros, E.~Shechtman, O.~Wang, The unreasonable effectiveness of deep features as a perceptual metric, in: Proceedings of the IEEE conference on computer vision and pattern recognition, 2018, pp. 586--595.

\bibitem{t-sne}
L.~Van~der Maaten, G.~Hinton, Visualizing data using t-sne., Journal of machine learning research 9~(11) (2008).

\bibitem{arcface}
J.~Deng, J.~Guo, X.~Niannan, S.~Zafeiriou, Arcface: Additive angular margin loss for deep face recognition, in: CVPR, 2019.

\bibitem{inception}
C.~Szegedy, S.~Ioffe, V.~Vanhoucke, A.~Alemi, Inception-v4, inception-resnet and the impact of residual connections on learning, in: Proceedings of the AAAI conference on artificial intelligence, Vol.~31, 2017.

\bibitem{DiT}
W.~Peebles, S.~Xie, Scalable diffusion models with transformers, in: Proceedings of the IEEE/CVF international conference on computer vision, 2023, pp. 4195--4205.

\end{thebibliography}




\end{document}